\documentclass[preprint,authoryear]{elsarticle}

\usepackage{amssymb,amsmath,amsfonts,bm,mathrsfs}
\usepackage{graphicx,subcaption,stfloats}
\graphicspath{{figures/}}
\usepackage{float}

\usepackage{cite,url,enumerate,xspace}
\usepackage{array,multirow,dcolumn,booktabs,longtable}
\usepackage{natbib}

\usepackage{lineno}
\usepackage[hmargin=1.25in, vmargin=1in]{geometry}
\usepackage{setspace}
\usepackage{xcolor}

\usepackage{hyperref}
\hypersetup{
  pdftitle={Fault Detection and Identification using Bayesian Recurrent Neural Networks},
  pdfborder={0 0 0}
}

\newcommand{\PreserveBackslash}[1]{\let\temp=\\#1\let\\=\temp}
\newcolumntype{C}[1]{>{\PreserveBackslash\centering}m{#1}}
\newcolumntype{d}[1]{D{.}{.}{#1}}

\newcommand{\x}{\+x}

\newcommand{\E}[2][]{\ensuremath{%
  \if\relax#1\relax\mathbb{E}\!\left[#2\right]
  \else\mathbb{E}_{#1}\left[#2\right]\fi}}

\newcommand{\R}{\ensuremath{\mathbb{R}}}

\newcommand{\figref}[1]{Figure~\ref{#1}}

\newcommand{\NOC}{NOC\xspace}

\makeatletter
\def\fps@figure{tbp}
\def\fps@table{tbp}
\makeatother

\usepackage{textcomp}

\journal{Computers \& Chemical Engineering}

\begin{document}

\begin{frontmatter}
\title{Fault Detection and Identification Using \\ Bayesian Recurrent Neural Networks}
\author[1]{Weike~Sun}
\author[2]{Antonio~R.~C.~Paiva\corref{cor1}}\ead{antonio.paiva@exxonmobil.com}
\author[2]{Peng~Xu}
\author[2]{Anantha~Sundaram}
\author[1]{Richard~D.~Braatz}
\cortext[cor1]{Corresponding author}
\address[1]{Department of Chemical Engineering, Massachusetts Institute of Technology, Cambridge, MA, USA}
\address[2]{Corporate Strategic Research, ExxonMobil Research and Engineering, Annandale, NJ, USA}

\begin{abstract}
In the processing and manufacturing industries, there has been a large push to produce higher quality products and ensure maximum efficiency of processes, which requires approaches to effectively detect and resolve disturbances to ensure optimal operations. While many types of disturbances can be compensated by a control system, it cannot handle some large process disruptions. As such, it is important to develop monitoring systems to effectively detect and identify those faults such that they can be quickly resolved by operators. This article proposes a novel probabilistic fault detection and identification method which adopts a newly developed deep learning approach using Bayesian recurrent neural networks (BRNNs) with variational dropout. The BRNN model is general and can model complex nonlinear dynamics. Moreover, compared to traditional statistic-based data-driven fault detection and identification methods, the proposed BRNN-based method yields uncertainty estimates which allow for simultaneous fault detection of chemical processes, direct fault identification, and fault propagation analysis. The performance of the method is demonstrated and contrasted to (dynamic) principal component analysis, which is widely applied in the industry, in the benchmark Tennessee Eastman process (TEP) and a real chemical manufacturing dataset.
\end{abstract}

%
%

\end{frontmatter}


\section{Introduction}\label{sec:intro}
In industrial manufacturing processes, a {\em fault} is defined as any abnormal deviation from the normal operating conditions (NOC). Faults are a concern because even small faults in a complex industrial system can initiate a series of events that result in loss of efficiency and reliability. As a result, there is a need for techniques to improve the process's reliability and up-time. Effective fault detection and identification (FDI) is important for monitoring components for making appropriate maintenance decisions. First, fault detection determines whether a fault has occurred in the system (also characterized as anomaly detection in other applications). Then fault identification determines which observation variables are most relevant to diagnosing the fault detected, thereby helping operators to focus on specific subsystems. Systems that can accurately and promptly detect and identify faults can more effectively inform operators and engineers and significantly reduce the effort and time to recover the system.

A number of FDI methods have been proposed in the literature. Since analytical and knowledge-based methods are impractical in most large-scale modern industrial processes, data-driven methods have dominated the literature for the past decade and have been effective in practice, taking advantage of increasing levels of instrumentation and widespread availability of sensor data \citep{Qin2009,Ge2013,Yin2014,Chiang2000}. The choice of model used to characterize the NOC and deviations thereof is still a crucial aspect in these methods because the limitations of the model lead to decreased detection rates or increased occurrence of fault alarms.

A number of data-driven methods including principal component analysis (PCA) \citep{Jolliffe2011}, partial least squares  \citep{Kourti1996}, Fisher discriminant analysis \citep{Fisher1936,Chiang2000}, and support vector machines \citep{Chiang2004}, 
have been applied for fault detection and identification in industry with varying degrees of success. The most widely used method is PCA which models the correlations between the process variables. PCA can detect faults effectively when the sensor measurements are highly correlated, which is often the case. For many processes, the temporal dynamics also need to be taken into consideration, especially when fast sampling rates are used, because the dynamics provide additional information through which to detect deviations from the NOC. To that end, DPCA has been proposed to handle serially correlated multivariate observations \citep{Ku1995}. DPCA can be viewed as a multivariate autoregressive model with exogenous inputs (ARX). PCA and DPCA are both limited by the linear model structure and correlations in the process' dynamics. Methods for extending PCA to nonlinearities such as kernel PCA and neural network-based PCA \citep{Lee20041,Hsieh2007} have been well studied only for static systems. As such, the development of approaches that can effectively model nonlinear system structure and dynamics has been an active research field.

Neural network (NN) based methods have also received significant attention because of their capability and flexibility for modeling complex structure and temporal dynamics. NN models have been used for fault detection in three general frameworks: (1)~as a fault classification tool between normal and known faulty conditions \citep{Zarei2014,Chine2016,Ince2016,Jia2016,Wu2018,Hu2018,Lee2017,Li2014,Zhang2017}, (2)~as a model of the input-output variable relationships during NOC \citep{Malhotra2015,Patan2008,Moustapha2008,Talebi2008,Wang2017,Nie2018}, or (3) as a generalization of the basic fault detection methods such as NN-based PCA \citep{Kramer1992,Dong1996} using statistical indices to monitor the process. The first two approaches are dominant for NN-based fault detection. The first approach can be highly effective due to the up-front knowledge of specific fault conditions to detect. It can also be set up to classify each fault which directly enables fault diagnosis. On the other hand, training these NNs requires substantial amounts of data under fault conditions but these data are usually quite limited for chemical manufacturing processes compared to NOC data. Moreover, it is hard to assess the performance of these methods for fault conditions other than for which the classifier is explicitly trained for.
In the second approach, NNs are used to model the process by capturing the nonlinear, multivariate, and temporal dependencies from inputs to outputs. In this approach, the NN models are typically trained on NOC data to predict the system outputs. This NN is then used for fault detection during runtime by comparing the predictions of the NN with the actual system output measurements, and a fault is detected if the difference is significantly large. This approach has the advantage that the NNs are trained using only NOC data, which is usually abundant, and that the NNs are not constrained by the type of fault because detection is marked from any significant deviation from the NOC. On the other hand, the model must accurately characterize these complex and potentially nonlinear structures between inputs and outputs in the process, or its fault detection will perform poorly as a result. Moreover, only faults that break the input-output relationships are considered, meaning that faults due to input disturbances will likely not be detected. 
In the third approach, NNs are used to account for nonlinearities in the process but, like other PCA-based methods, fail to appropriately model the process dynamics.

There are other challenges regarding both approaches, which have limited the application of NNs in industrial process monitoring. First, fault identification has not yet been properly addressed. Once a fault is detected, it is typically difficult to identify the input variables most relevant to the fault from a complex NN model. Even if an NN is trained to directly classify the fault, the underlying cause may still be unclear if there are multiple explanations for the observed fault type. Secondly, standard NNs are deterministic models which lack an estimate of the uncertainty in the model outputs. However, uncertainty and probabilistic estimates are important to assess the confidence level associated with the decision of detecting a fault and for fault identification. Lastly, NNs are prone to overfitting, meaning that they `memorize' the particular characteristics of the training data that are not relevant for new data. This overfitting must be addressed to ensure good generalization to the full space of operating conditions of a complex industrial process.

For fault identification, contribution plots \citep{Miller1998} are one of the most popular techniques for providing information on the variables that are most strongly related to the faults. In the context of PCA-based methods, contribution plots are obtained by quantifying the contribution of each process variable to the individual scores of the PCA representation \citep{Westerhuis2000}. Methods based on the contribution of each process variable in the residual space have also been developed \citep{Wise1989}. However, the aforementioned limitations of PCA-based approaches will also be reflected in the identification procedure and those methods require extra processing steps after fault detection. Moreover, those methods only provide the relative contribution value of each variable which is not very useful. A more valuable and precise measure to aid operators in diagnosis would be the probable severity of each affected variable. On the other hand, it has been hard to extend contributions plots to NNs due to the complex and nonlinear relationships between predictions and model inputs.

This article proposes a novel end-to-end FDI framework, which adopts a recently developed Bayesian recurrent neural network (BRNN) architecture \citep{Gal2016theoretically}. The proposed FDI framework is fundamentally different from the two types of frameworks that have been previously used in the NN-based fault detection literature. The proposed framework uses BRNNs to model the joint distribution and dynamics between all process variables. This framework provides estimates of the prediction uncertainty, which capture both model uncertainty and the inherent noise in the data. The BRNN is realized using the variational dropout approach proposed in \citep{Gal2016dropout,Gal2016theoretically} due to its simplicity, regularization capability, strong generalization ability, and scalability. 

To the best of our knowledge, this work is the first time that Bayesian spatio-temporal models, and BRNNs in particular, have been successfully applied to FDI in the chemical manufacturing industry. The proposed approach tackles three key challenges typical of manufacturing systems: (1)~nonlinearity, (2)~non-Gaussian distributed variables, and (3)~high degree of spatio-temporal correlations (i.e., temporal and sensor correlations). Furthermore, the probabilistic framework provided by BRNNs enables more sensitive and robust FDI. Fault identification through the proposed BRNN-based approach provides easily interpretable visualizations to the plant operators, for quick fault type categorization, analysis of the possible fault propagation path, and root cause determination using engineering judgment.

The remainder of this paper is organized as follows. Section~\ref{section:BRNN2} provides a brief introduction to RNNs and BRNNs, and describes the variational dropout approach used in this paper for inference in BRNNs. Section~\ref{sec:methodology-for-fault-detection-and-identification} presents the proposed BRNN-based FDI methodology. In Section~\ref{sec:case-studies}, the effectiveness of the proposed approach is demonstrated and compared to (D)PCA-based methods in the Tennessee Eastman process and a real chemical manufacturing process, followed by the conclusion in Section~\ref{sec:conclusion}.

\section{Background}\label{section:BRNN2}

\subsection{Recurrent Neural Networks}\label{sec:rnns}

\begin{figure}
  \centering
  \includegraphics[width=0.48\textwidth]{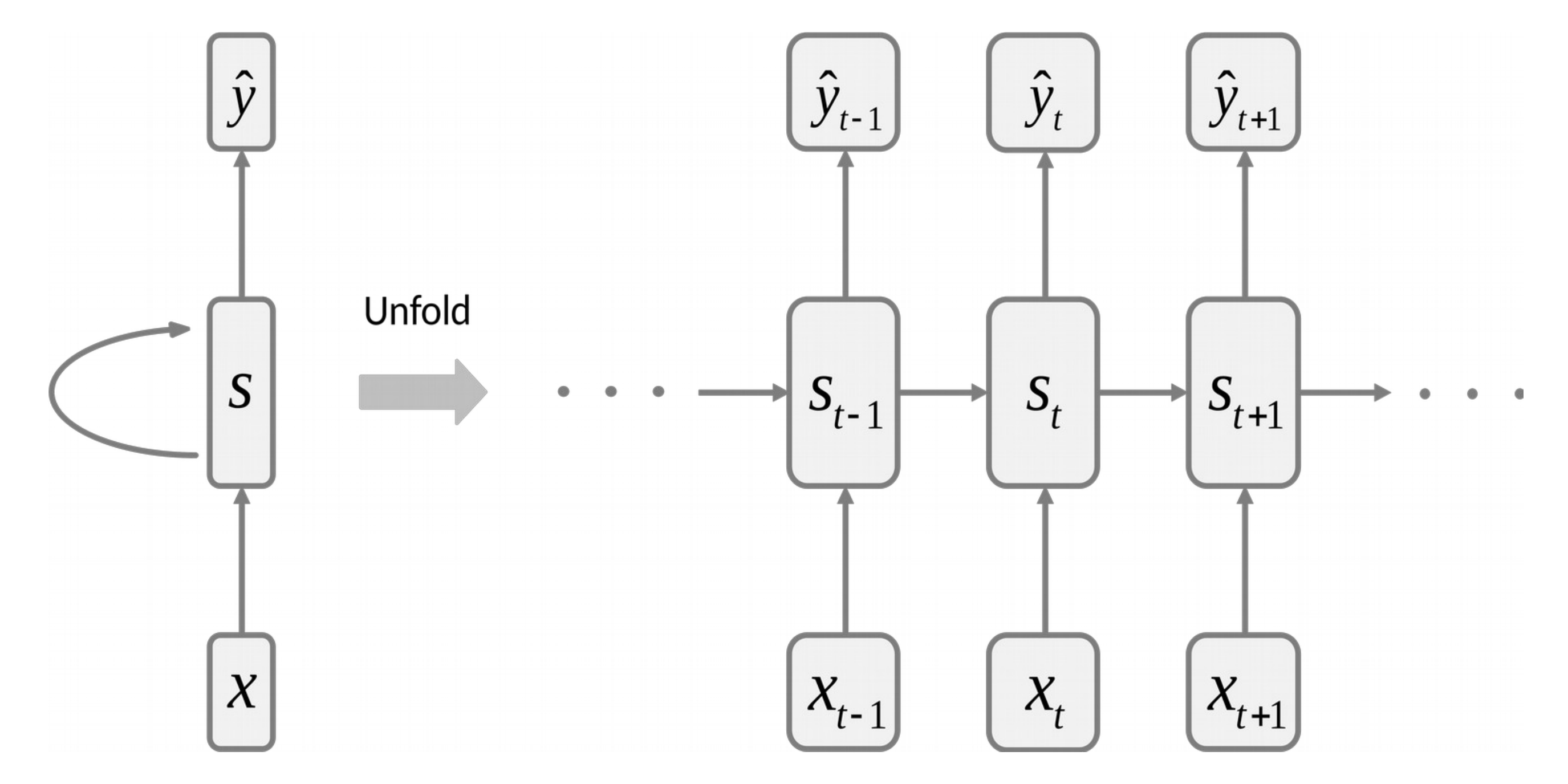}
  \caption{A simple RNN structure with one recurrent layer and showing the unfolding in time of the sequence of its forward computation. The RNN includes the input variable $\bm{x}_t$, state variable $\bm{s}_t$ and outputs $\hat{\bm{y}}_t$. The state variable $\bm{s}_t$ is calculated based on the previous state $\bm{s}_{t-1}$ and the current input $\bm{x}_t$. The RNN output $\hat{\bm{y}}_t$ is then calculated based on the current state. In this way, the input sequence $\bm{x}_t$ is mapped to output sequence $\hat{\bm{y}}_t$ with each $\hat{\bm{y}}_t$ depending on all previous inputs. The model parameters $\omega = \{\bm{W}_s, \bm{U}_s, \bm{W}_s, \bm{b}_s, \bm{b}_y\}$ are shared at each time step.}
  \label{arm:BRNN1}
\end{figure}

RNNs were developed in the 1980s~\citep{Rumelhart1986}. Since then, RNNs have been shown to achieve state-of-the-art performance on a wide range of sequential data modeling tasks, including language modeling, speech recognition, image captioning, and music composition \citep{Wu2016,Jozefowicz2016,Merity2016}. Generally speaking, an RNN comprises an input layer, one or more hidden recurrent layers, and an output layer. The input layer corresponds directly to the input data, and hidden recurrent layers capture the state with the response of its nodes being added to the inputs on the next time step. At each time $t$, denote the input to the network as $\bm{x}_t\in\R^{m_x}$, the state (i.e., output of the hidden layer) as $\bm{s}_t\in\R^{m_s}$, and the RNN output as $\hat{\bm{y}}_t\in\R^{m_y}$. They are represented as row vectors in the equations.
Accordingly, the state and output layers have the general form
\begin{equation}\label{eq:rnn}
\begin{split}
  \bm{s}_t &= f_s\!\left(\bm{x}_t, \bm{s}_{t-1} \right | \theta_s) \\
  \hat{\bm{y}}_t &= \bm{W}_y\bm{s}_t  + \bm{b}_y \\
\end{split}
\end{equation}
where the subscript $s=1,\dots,m_s$ is the index over hidden layer nodes, $\theta_s$ and $f_s$ denote the corresponding hidden layer parameter/weights and nonlinear operator for each node, and $\bm{W}_y\in\R^{m_y \times m_s}$ and $\bm{b}_y\in\R^{m_y}$ are the output layer parameters. The new state of the network depends on its value at the previous time step, emblematic of recurrent architectures. This dependency, and the unfolding through time, is depicted in Figure~\ref{arm:BRNN1}. A linear output layer is commonly used for regression tasks.

In the simpler form of nodes, the state is computed as \citep{Elman1990}
\begin{equation}\label{eq:rnn:state.vanilla}
  \bm{s}_t = \phi\!\left(\bm{W}_s\bm{x}_t + \bm{U}_s\bm{s}_{t-1} + \bm{b}_s \right)
\end{equation}
where $\bm{W}_s\in\R^{m_s \times m_x}$, $\bm{U}_s\in\R^{m_s \times m_s}$, and $\bm{b}_s\in\R^{m_s}$ are the hidden layer parameters (denoted $\theta$ above), and $\phi$ is an element-wise activation function such as the logistic, hyperbolic tangent, or a rectifier linear function.

As can be explicitly observed from the mathematical formulation in Equation~\ref{eq:rnn}, RNNs are essentially state-space models capable of modeling nonlinear dependencies. RNNs can capture complex nonlinear dynamics of a system in the state. Also, by appropriately training the parameters, RNNs can adapt to the right level of temporal depth. Thus, RNN models are considerably more powerful for modeling complex industrial processes in comparison to traditional statistical methods.

It is worth noting that different RNN architectures have been proposed \citep{Jordan1997}, with the formulation in Equation~\ref{eq:rnn} corresponding to Elman's architecture \citep{Elman1990}, which has been widely used in the recent deep learning RNN implementations and applications.

In order to learn the parameters of the RNN, an optimization problem is defined with regard to an appropriate loss function. For regression tasks, the loss function is typically chosen to be the mean squared loss,
\begin{equation}
  J(\Theta) = \frac{1}{N}\sum_{t = 1}^{N}\Vert\bm{y}_t - \hat{\bm{y}}_t\Vert^2_2,
\end{equation}
or the cross-entropy loss for classification purposes,
\begin{equation}
  J(\Theta) = - \sum_{t=1}^{N}\bm{y}_t\log \hat{\bm{y}}_t
\end{equation}
where $\Theta$ denotes the collection of all RNN model parameters, and $\bm{y}_t$ is the desired output at time step $t$. In addition, $L_2$ regularization terms are often added to help prevent overfitting, resulting in the overall minimization objective
\begin{equation}
  L(\Theta) = J(\Theta) + \lambda\left( \Vert{\bm{W}_s}\Vert^2 + \Vert\bm{W}_y\Vert^2 + \Vert{\bm{U}_s}\Vert^2 \right)
\end{equation}
where $\lambda$ is the regularization (aka weight decay) parameter.

Because the recurrence introduces dependencies between time steps, training RNNs involves backpropagation through time~(BPTT) to compute the gradient update of the model parameters that minimizes the loss function~\citep{Werbos1974,Werbos1990}. BPTT corresponds to an unfolding of the network over a number of time steps, as depicted in Figure~\ref{arm:BRNN1}. For BPTT, the difference between network outputs and target values is first calculated and stored for each time step in a forward pass, and then the weight gradient updates are calculated as the network is ``rolled back''. However, simple RNNs trained with BPTT can have difficulties learning long-range time dependencies due to the vanishing gradient problem \citep{Bengio1994}. To alleviate the vanishing gradient problem, recurrent node gating mechanisms have been recently developed. These gating mechanisms allow information and the gradients to flow through the unrolled network with minimal attenuation if determined to be necessary by BPTT. These gating mechanisms resulted in two popular variations on RNN hidden units: LSTM units \citep{Hochreiter1997} and GRUs \citep{Cho2014}. RNNs with LSTM and GRU units have been reported to show salient performance \citep{Graves2013,Cakir2015}.

\subsection{BRNNs}\label{sec:brnns}
BRNNs combine statistical modeling of RNN parameters to obtain a probabilistic model of input-output mapping. As such, instead of point estimates, BRNNs can effectively perform Bayesian inference which provides probabilistic distributions over the outputs.

To realize that capability, BRNNs view the model parameters $\omega = \{\bm{W}_s,\bm{W}_y,\bm{U}_s,\bm{b}_s,\bm{b}_y \}$ as random variables from a prior distribution $p(\omega)$. Expressing the functional dependence in Equation~\ref{eq:rnn} as $\bm{s}_t = \bm{f}_s^{\omega}(\bm{x}_t,\bm{s}_{t - 1})$ and $\hat{\bm{y}}_t = \bm{f}_y^{\omega}\left( \bm{s}_t \right)$, the likelihood of the output for each data point is
\begin{equation}
  p(\bm{y}_t | \omega, \bm{x}_t, \bm{s}_t, \tau)
  = N\!\left(\bm{y}_t | \bm{f}_{y}^{\omega}\left(\bm{f}_{s}^{\omega}\left( \bm{x}_t, \bm{s}_{t-1} \right),
      \tau^{-1}\bm{I}_{D} \right) \right)
  \label{eq:training.posterior}
\end{equation}
where $\tau$ is the precision parameter that reflects the intrinsic noise in the data, and the likelihood function is assumed to have a normal distribution for simplicity. Note how the likelihood function is evaluated with respect to forward passes through the NN.

Then, given a training dataset comprising $\bm{X}$ and $\bm{Y}$, learning entails estimating the posterior distribution $p(\omega|\bm{X},\bm{Y})$ over the space of parameters. With the updated distribution, the distribution of a predicted output $\bm{y}^*$ can be obtained by integration
\begin{equation}
  p(\bm{y}^* | \bm{x}^*, \bm{X}, \bm{Y})
  = \int p(\bm{y}^* | \bm{x}^*, \omega) p(\omega|\bm{X},\bm{Y}) d\omega
  \label{eq:output.posterior}
\end{equation}
where the dependency on the precision parameter, state, and past inputs are not shown to simplify the expression. 
For the prior distribution, standard zero-mean Gaussian priors over the weight matrices $p(\bm{W})$ and $p(\bm{U})$ are typically chosen, with point estimates for the bias vectors assumed for simplicity.
The uncertainty in the prediction will be directly reflected in the posterior distribution $p(\bm{y}^* | \bm{x}^*, \bm{X}, \bm{Y})$.

In complex models such as NNs, however, the exact inference of the posterior is not possible. Moreover, traditional algorithms for approximating the Bayesian inference are generally not applicable to train RNNs having a large number of parameters or complex architectures.
To overcome this limitation, several approximation inference methods have been proposed, including variational dropout \citep{Gal2016theoretically,Gal2016dropout}, Bayes by BackProp \citep{Pawlowski2017, Fortunato2017}, multiplicative normalizing flows \citep{Louizos2017}, and probabilistic backpropagation \citep{Hernandez2015}. Among all those techniques, the variational dropout technique proposed by  \citet{Gal2016theoretically} is adopted in this paper because of its simplicity and generalization capability. 
In particular, variational dropout can be applied to any RNN architecture without modification on the underlying NN structure, and only concurrent runs of the trained model are needed for online application. Details of this algorithm are reviewed in the next section.

\subsection{Variational Dropout as Bayesian Approximation}\label{sec:variational-dropout-as-bayesian-approximation}

\citet{Gal2016dropout} showed how dropout could be used as a general variational approximation to the posterior of Bayesian neural networks (BNNs), which can be applied directly to a variety of NN architectures. The main advantage of this `variational dropout' approach is that it does not require significant modifications to the model architecture and training method, unlike other probabilistic approximation methods. Moreover, the uncertainty estimation incurs only the computation cost due to multiple stochastic forward passes through the network to generate samples of the posterior distribution.

Therefore, variational dropout is used here as a variational inference approach for BNNs. Variational inference is a technique used to approximate an intractable posterior distribution $p(\omega|\bm{X}, \bm{Y})$ with a simpler parameterized distribution $q(\omega)$. Then, the integration in Equation \ref{eq:output.posterior} can be approximated simply by MC integration using $q(\omega)$. Specifically, the approximation distribution is factorized over the weight matrices in $\omega$. For each row $\bm{w}_k$, variational dropout imposes a variation distribution comprising a mixture of two Gaussian distributions with small variances,
\begin{equation}
  q(\bm{w}_k)
    = p N(\bm{w}_k | 0, \sigma^2\bm{I})
    + (1-p)N(\bm{w}_k | \bm{m}_k,\sigma^2\bm{I}),
\end{equation}
where $p$ is the predefined dropout probability, $\sigma^2$ is a small precision parameter, and $\bm{m}_k$ is a variational parameter. The learning problem is then casted into an optimization problem by minimizing the KLD between $q(\omega)$ and $p(\omega | \bm{X}, \bm{Y})$. It can be shown that optimizing the loss function using dropout is equivalent to minimizing KL$( q(\omega) \| p(\omega | \bm{X}, \bm{Y})))$ \citep{Gal2016dropout}, which updates the variational parameter. Although variational inference is a biased approximation, it has been shown to work well in practice.

\begin{figure}
  \centering
  \includegraphics[width=11cm]{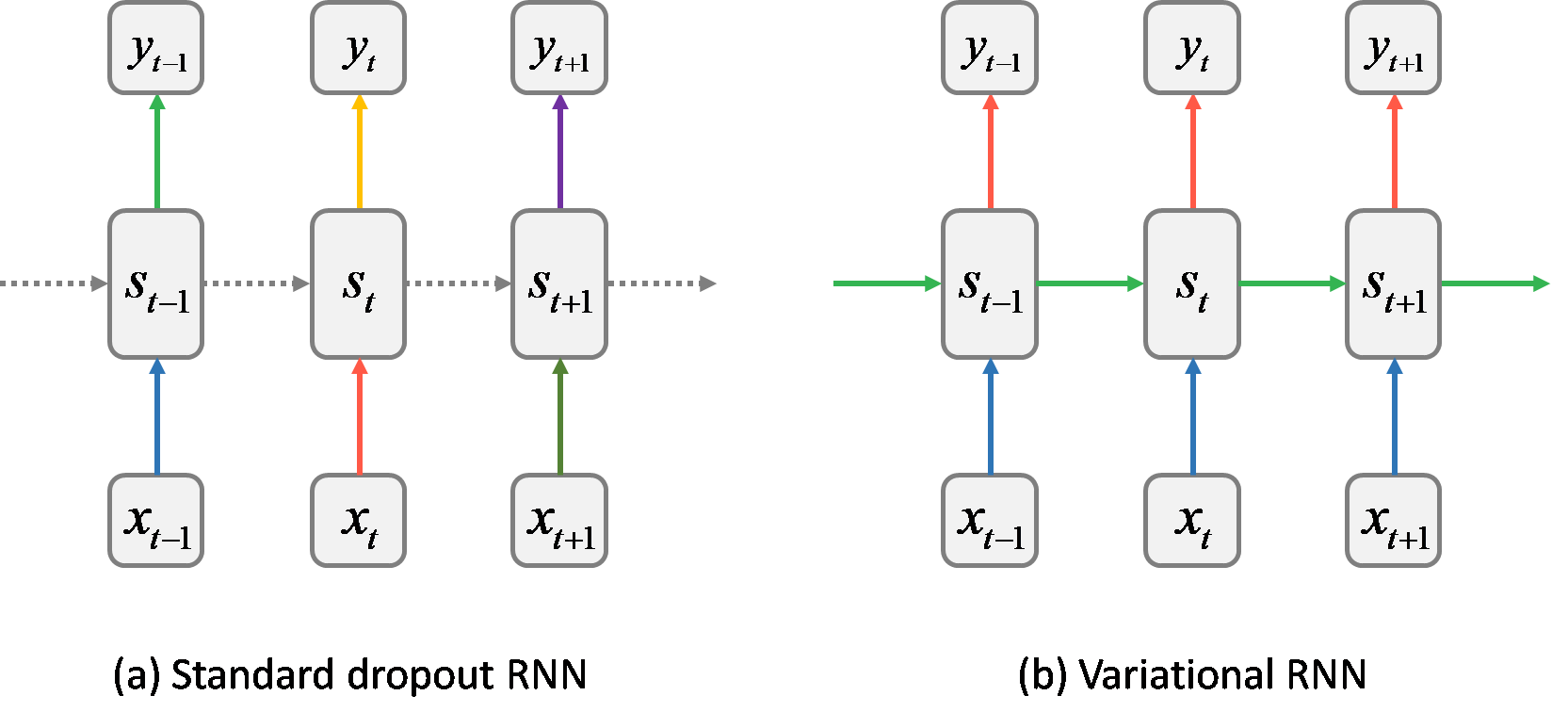}
  \caption{Illustration of the variational dropout technique (right) compared to standard dropout technique (left) for a simple RNN. Each graph shows units unfolded over time, with the lower level for inputs, middle level for state units, and upper level for output units. Vertical arrows represent the connections from inputs to outputs while horizontal arrows represent recurrent connections. The arrows with dashed grey lines represent the standard connection without dropout. Colored lines represent dropout connections with different colors for different dropout masks. (Left) In the standard dropout technique, no dropout is applied for the recurrent layers, while other connections have different dropout masks at different time steps. (Right) For the variational dropout approach proposed in \citep{Gal2016theoretically}, dropout is applied to both input, recurrent, and output layers with the same dropout mask at different time steps. Variational dropout is applied during both training and testing.}
  \label{fig:variationalRNN}
\end{figure}

Variational dropout requires caution when applied in the context of RNNs, however. Because of the recurrence, na\"ively applying standard dropout \citep{Srivastava2014} with different masks at each time step of an RNN can lead to model instabilities and disrupt an RNN's capability to model a sequence \citep{Pham2014,Pachitariu2013}. We use the approach in \citep{Gal2016theoretically} to resolve these issues. Under these circumstances, variational dropout has been shown to also act as an effective regularization method for reducing overfitting by preventing co-adaptions in RNNs \citep{Gal2016theoretically}.

The implementation of BRNNs with variational dropout is relatively straightforward. During both training and testing, the variational approximation involves sampling the model distribution with regard to the variational distribution over the weights, which is implemented by dropping out (i.e., forcing to zero) randomly chosen inputs, outputs, and hidden states. This step results in multiple random realizations of the RNN model, each obtained by implicitly removing a portion of the inputs, outputs, or hidden states. However, as detailed in \citep{Gal2016theoretically}, it is crucial for RNNs that the dropout mask used for each model realization be kept fixed between time steps. In other words, the dropout mask of which elements are zeroed out is sampled and frozen for each time sequence sample. This sampling characteristic is contrasted to standard dropout in Figure \ref{fig:variationalRNN}.

Variational dropout applied during testing can be viewed as an approximation to MC samples from the posterior predictive distribution, $p\left( \omega | \bm{X},\bm{Y} \right)$. Given a new observation $\bm{x}^*$, by forward passing it $N$ times,  $N$ samples $\left\{ {\hat{\bm{y}}}^*(i) \right\}_{i = 1,\dots,N}$ are collected of the approximate predictive posterior. The corresponding empirical estimators for the posterior predictive mean, standard deviation, and covariance are
\begin{gather}
  E(\hat{\bm{y}}^*) \approx \frac{1}{N}\sum_{i = 1}^{N}\hat{\bm{y}}^*(i)
  \\
  \text{std}(\hat{y}^*) \approx \sqrt{\tau^{-1} + \frac{1}{N}\sum_{i=1}^{N} \left(\hat{y}^*(i)\right)^2 - E(\hat{y}^*)^2}
  \\
  \text{cov}( \hat{\bm{y}}^*) \approx \tau^{-1}\bm{I}_{D} + \frac{1}{N}\sum_{i=1}^{N}{{\hat{\bm{y}}^*(i)}^\top \hat{\bm{y}}^*(i)} - E(\hat{\bm{y}}^*)^\top E(\hat{\bm{y}}^*)
\end{gather} 
where $\tau$ can be estimated as $\tau = \frac{pl^2}{2N\lambda}$ given a predefined regularization/weight-decay parameter $\lambda$, and prior length scale $l$ \citep{Gal2016dropout}. Higher order statistics can also be estimated using the samples by moment-matching.

Since the forward passes involve simply a number of independent and fixed realizations of the RNN model distribution, they can be done concurrently, thus making variational dropout a good candidate for online monitoring. In the next section, the proposed novel FDI scheme is explained in detail. While this methodology is described here in the context of chemical process monitoring, it can be observed that it could be readily extended to other manufacturing industries.

\section{Methodology for FDI}\label{sec:methodology-for-fault-detection-and-identification}

\begin{figure}
  \centering
  \includegraphics[width=0.45\textwidth]{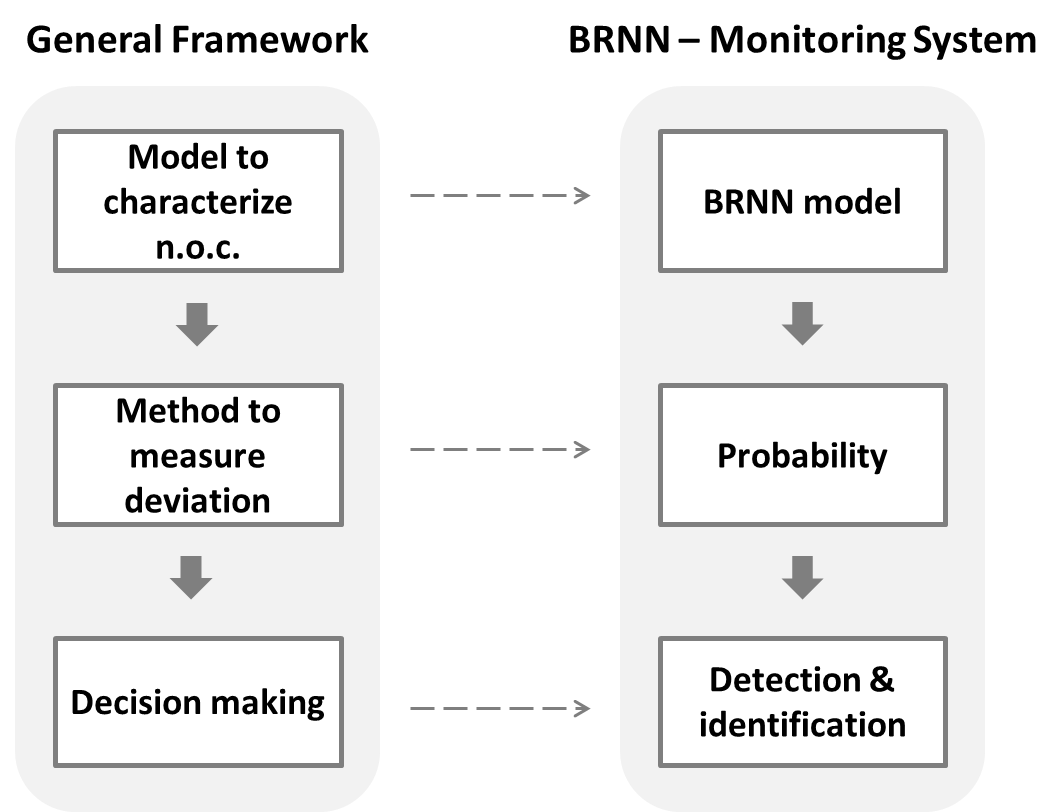}
  \caption{General procedure for process monitoring system development (left) versus procedure for developing BRNN-based FDI system (right). The general framework to establish a monitoring system begins with a model to characterize NOC behavior, such as using the BRNN model to learn the NOC pattern from the training data. Then, the method to measure the deviation of a particular observation to the NOC region is chosen. In our case, the process observations are compared to the BRNN posterior predictive distributions. Finally, the decision will involve determining whether the acquired observation is from the NOC or not (i.e., compare deviations of observations for fault detection and assess which variables significantly deviate from the NOC for identification).}
  \label{fig:detection.process}
\end{figure}

The design of a FDI system generally begins with the development of a model to characterize the normal operating characteristics of a process. Historical data collected during the NOC are used to build the model, which means this learning problem is unsupervised. Then, an approach must be established to characterize the magnitude of the deviation from the NOC based on the developed model and to determine when deviations are considered to be outside of the NOC. For example, the $T^2$ and $Q$ statistics are commonly used to measure the distance of the observation to the NOC region in PCA-based models and thresholds thereon \citep{Chiang2000}. Finally, given a new observation $\bm{x}^*$, these measures are calculated to determine whether $\bm{x}^*$ deviates substantially from the NOC (fault detection) and, if that is the case, which variables are significantly affected (fault identification), thereby assisting in locating and troubleshooting the fault.

Specifically, this paper proposes using a BRNN with variational dropout to build the probabilistic model, denoted as $f^{\omega}(\cdot)$, and characterize the NOC and its intrinsic variability. As discussed earlier, BRNNs are capable of extracting the nonlinear spatial and temporal signatures in the data that are critical for characterizing complex chemical processes. Moreover, BRNNs provide probabilistic estimates of the likelihood of the observations with regard to its inferred posterior distribution of the variable values. These likelihood estimates lend themselves to be used to assess the current deviation level from the NOC region. Accordingly, observations are detected as faults whenever their deviation is above a threshold, determined such that the number of false alarms under the NOC does not exceed a predefined level. Fault identification then involves determining which process attributes are deviating significantly. This general framework for BRNN-based FDI is summarized in Figure \ref{fig:detection.process} and described in detail in the next sections.

\subsection{Fault Detection}\label{fault-detection}

The first step toward fault detection is to learn a model to characterizing the NOC. This step involves training a BRNN with variational dropout to model the dynamics in time, correlations between sensors, and the prediction uncertainty resulting from model mismatch and inherent system variability/noise.

\begin{figure}
  \centering
  \includegraphics[width=0.44\textwidth]{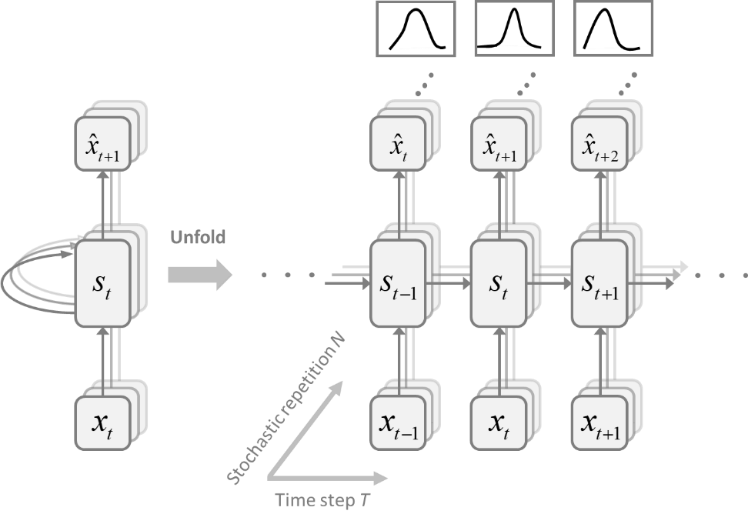}
  \caption{Depiction of BRNN model using variational dropout (left) for
    FDI. The BRNN model uses the current
    observation and state to predict the next system observation. The BRNN
    model is unrolled in two dimensions (right): the time of the computation
    involved in its forward computation and the stochastic repetition by
    variational dropout. At each time step, stochastic variational dropout
    is applied $N$ times and the corresponding MC prediction samples
    $\left\{ \hat{\bm{x}}_t(i) \right\}_{i = 1,\dots,N}$
    are used to approximate the posterior predictive distribution for
    that time step. For the next time step, the same procedure is repeated
    and MC samples
    $\left\{ \hat{\bm{x}}_{t + 1}(i) \right\}_{i = 1,\dots,N}$
    are collected and used to approximate the distribution.}
  \label{fig:brnn.MCdropout}
\end{figure}

The BRNN model is trained directly on historical NOC data. Specifically, this step involves setting a training problem wherein the BRNN model uses the past context (as captured by its state) and current observation to predict the next observation, as depicted in Figure \ref{fig:brnn.MCdropout}. During training, BPTT is applied to batches of time subsequences with one variational dropout mask sampled per sequence, as explained in Section \ref{sec:variational-dropout-as-bayesian-approximation}.

After training, the model output $\hat{\bm{x}}_{t+1}$ from the BRNN is sampled from the posterior predictive distribution for next observation $\bm{x}_{t+1}$ via variational dropout model realizations. That is, at each time step $t$, the stochastic forward pass is repeated $N$ times, each with a different dropout mask, and the predictive distribution of the output for $t+1$ is approximated based on the MC samples of the BRNN model, $\left\{ \hat{\bm{x}}_{t+1}(i) \right\}_{i = 1,\dots,N}$. Then, when the true observation $\bm{x}_{t+1}$ is available, it is compared to the predictive distribution and deemed as abnormal if it significantly deviates from the predictive distribution. Finally, the true observation is fed into the BRNN model and the procedure is repeated for the next time step.

Notice that the predictive distribution is evolving, which provides an adaptive decision boundary for the next measurement. This adaptive decision boundary is calculated based on all the useful past system information, which takes into consideration both the spatial and temporal correlations in the data. Further combined with the potential ability to model nonlinear correlations, this property increases both the detection sensitivity and robustness because of the increasing accuracy in modeling NOC pattern.

Depending on the complexity of the system and observed properties of the predictive distribution from the BRNN model, below is a description of two methodologies to quantify the deviation magnitude of each observation to its corresponding predictive distribution. The first method is faster and simpler to implement, but is limited to Gaussian predictive posterior distributions. The second method approximates the posterior distribution non-parametrically and is much more flexible, but requires tuning an additional density estimation parameter.

\subsubsection{Method 1: Squared Mahalanobis distance for Gaussian predictive distributions}

If the predictive distribution is Gaussian, or well approximated as such, the squared Mahalanobis distance can be used to characterize the magnitude of the deviation. First, the MC samples at time $t$ of the predictive distribution, $\left\{\hat{\bm{x}}_t(i) \right\}_{i = 1,\dots,N}$, are used to approximate the sample mean $\bm{\mu}_t$ and covariance $\bm{S}_t$:
\begin{gather}
  \bm{\mu}_t \approx \frac{1}{N}\sum_{i = 1}^{N}\hat{\bm{x}}_{t(i)}
  \\
  \bm{S}_t \approx \tau^{-1}\bm{I}_D
      + \frac{1}{N}\sum_{i = 1}^{N}{{\hat{\bm{x}}_t(i)}^\top {\hat{\bm{x}}}_t(i)}
      - \bm{\mu}_t^\top \bm{\mu}_t.
\end{gather}
Then, when the true observation $\bm{x}_t$ is available, the squared Mahalanobis distance is calculated as
\begin{equation}
M^2 = (\bm{x}_t - \bm{\mu}_t)^\top \bm{S}_t^{-1} (\bm{x}_t - \bm{\mu}_t).
\end{equation}

A larger value of $M^2$ indicates that the observation $\bm{x}_t$ is far away from the predicted mean and there is a higher likelihood that it corresponds to a fault. The detection threshold $M_{\text{th}}^2$ is determined with regard to a chosen maximum FAR $\alpha$ on a validation dataset. That is, the threshold is the $100(1-\alpha)^{\mathrm{th}}$ percentile of the $M^2$ statistic in the MC samples of the validation dataset. Therefore, any data point with $M^2$ exceeding the threshold ($M^2 > M_{\text{th}}^2)$ should be detected as a fault.

\subsubsection{Method 2: Local density ratio (LDR) for non-Gaussian predictive distributions}

If the predictive distribution is not well characterized by a Gaussian distribution (e.g., is multi-modal), then non-parametric methods are necessary to quantify the abnormality of each observation. For those cases, a LDR method is proposed, which is closely related to the so-called local outlier factor \citep{Breunig2000}. The LDR statistic quantifies the abnormality of each new observation with respect to its predictive distribution using an estimate of the density around the observation based on its $k$NN.

The $k$NN local density estimate $\hat{f}\left( \bm{x} \right)$ can be calculated as \citep{Duda2001}
\begin{equation}
  \hat{f}(\bm{x}) = \frac{k}{\sum_{p \in N_k(\bm{x})} d(p,\bm{x})}
\end{equation}
where $N_k(\bm{x})$ denotes the set of $k$NN of $\bm{x}$ in $\left\{ {\hat{\bm{x}}}_t(i) \right\}_{i=1,\dots,N}$ and $d(p,\bm{x})$ is the Euclidean distance between $\bm{x}$ and a point $p \in N_k(\bm{x})$. Intuitively, the points close to its $k$NN will have high local density values, whereas points in more sparsely sampled or spread out areas will have low density.

Then, the LDR for an observation $\bm{x}_t$ is defined as
\begin{equation}
  \text{LDR}(\bm{x}_t) = \frac{\frac{1}{k}\sum_{p \in N_k(\bm{x}_t)} \hat{f}(p)}{\hat{f}(\bm{x}_t)}
\end{equation}
which is the ratio of the averaged local density of the $k$NN of $\bm{x}_t$ in $\left\{ {\hat{\bm{x}}}_t(i) \right\}_{i = 1,\dots,N}$ to the local density of $\bm{x}_t$. A larger value of the LDR$(\bm{x}_t)$ means that the observed point is far away from the samples of the prediction posterior and thus indicates higher likelihood that the observation $\bm{x}_t$ is abnormal.

The number of $k$NN specifies the smallest number of data points in a cluster that will be considered as abnormal and is crucial for the algorithm to perform properly. In general, this number defines a tradeoff, because a small value of $k$ will result in large fluctuations, whereas a very large value of $k$ will reduce the detection sensitivity. As recommended in \citep{Breunig2000}, a minimum and maximum $k$ can be chosen and, for each observation, the final value can be set equal to the maximum of LDR over $k$. The detection threshold $\text{LDR}_{\text{th}}$ is obtained similarly to $M_{\text{th}}^2$.

\subsection{Fault Identification}\label{fault-identification}

Once a fault is detected, the next goal is to identify the main variables associated with the fault. Without using labeled fault examples, this step involves determining the observation variables with the abnormal deviations, which are most relevant to locate and troubleshoot the fault.

BRNN fault identification is obtained by applying the fault detection approach but independently for each variable. To determine which variables deviate abnormally, each observation variable is compared to its corresponding predicted marginal posterior distribution estimated from the BRNN samples. More specifically, the observation $x_t^j$, corresponding to the $j^{\mathrm{th}}$ system variable at time $t$, is compared to the predictive posterior distribution characterized by the samples $\{\hat{x}_t^j(i)\}_{i = 1,\dots,N}$. This variable-wise comparison allows the identification of variables with values in low probability areas and thus more likely to be relevant for diagnosing the fault.

The marginal posterior distributions used for identification still take into consideration the spatial and temporal correlations in past data observations. Hence, the marginal distribution for each variable also evolves with dynamics that depend on other variables and past observations. This analysis sacrifices some information with regard to the complete joint distribution, as considered during fault detection, but is necessary to obtain variable specificity.

As with fault detection, two methodologies to quantify the fault identification deviation are described here, depending on the properties or assumptions placed on the predictive distribution. The same considerations apply to these methodologies.

\subsubsection{Method 1: Standard deviation for Gaussian predictive distributions}

Assuming that the predictive distribution can be approximated by a Gaussian distribution, the number of standard deviations of each variable to its predictive mean can be used to measure the deviation. Using the same MC samples of the posterior predictive distribution generated for fault detection at time $t$, $\left\{ \hat{\bm{x}}_t(i) = \{\hat{{x}}_t^l(i)\}_{l = 1,\ldots,m_x} \right\}_{i = 1,\dots,N}$, the mean $\mu_t^l$ and standard deviation $\sigma_t^l$ of each variable ${\hat{x}}_t^l$ can be estimated by
\begin{gather}
  \mu_t^l \approx \frac{1}{N} \sum_{i = 1}^{N} \hat{x}_t^l(i)
  \\
  \sigma_t^l \approx \sqrt{\tau^{-1} + \frac{1}{N}\sum_{i = 1}^{N} \left(\hat{x}_t^l(i)\right)^2 - (\mu_t^l)^2}
\end{gather}

The deviation for each variable $D^l$, $l\in\{1,\dots,m_x\}$, is then calculated from
\begin{equation}
  D^l = \frac{x_t^l - \mu_t^l}{\sigma_t^l}
  \label{eq:fault.id:Dl.stat}
\end{equation}
The $D^l$ can be either negative or positive, unlike $M^2$ which can only be positive. Under a Gaussian approximation, variables are identified as significantly affected by the disturbance based only on whether $D^l$ has a large magnitude (i.e., absolute value). Still, the sign of the deviation (positive or negative) can be helpful to operators because the sign explicitly indicates whether the variable is significantly higher or lower than expected. For a predefined significance level, the NOC validation dataset can be used to determine thresholds $\{D_{\text{th}}^l\}_{i = 1,\dots,N}$ such that variables with ($D^l >$ $D_{\text{th}}^l$) are explicitly highlighted as abnormal.

\subsubsection{Method 2: LDR for non-Gaussian predictive distributions}

For more general distributions, and similarly to the fault detection procedure, the LDR can be used element-wise for fault identification by considering each variable separately in the calculation of the LDR. Given the true measurement $\bm{x}_t = \left\{ x_t^l \right\}_{l = 1,\dots,m_x}$ and MC samples from predictive distribution $\left\{ {\{\hat{x}_t^l(i)\}}_{l = 1,\dots,m_x} \right\}_{i = 1,\dots,N}$, the LDR for variable $l$ can be calculated by
\begin{gather}
  \hat{f}(x_t^l) = \frac{k}{\sum_{p^l \in N_k(x_t^l)} d(p^l, x_t^l)}
  \\
  \text{LDR}^l(x_t^l) = \frac{\frac{1}{k}\sum_{p^l \in N_k(x_t^l)} \hat{f}(p^l)}{\hat{f}(x_t^l)}
\end{gather}
where $p^l$ is one of the $k$NN of $x_t^l$ and $d(p^l, x_t^l)$ is the Euclidean distance between the $p^l$ and $x_t^l$ sample. The same rule for selecting the number of nearest neighbors $k$ discussed with regard to fault detection can be used here.

The variables associated with a large value of $\text{LDR}^l$ can be explicitly selected as significantly affected by the fault. This is done similarly as for fault detection using the NOC validation dataset to determine a threshold $\text{LDR}_{\text{th}}^l$ above which variables are considered abnormal. Alternatively, the variables can simply be sorted from the largest to the smallest such to emphasize the system variables that deviate the most.

\subsubsection{Fault identification plots}\label{sec:fault.idplots}

The fault identification statistics of each variable can be visualized by plotting their values over time. The resulting plots are visually similar to contribution plots \citep{Miller1998,Zhu2014}. Their interpretation and analysis, however, are fundamentally different and are referred to here as {\em identification plots}. The main distinction is that the statistics in identification plots are specific to the current status of each variable and its dynamics, rather than as a relative component of a global statistic. These plots provide greater specificity in the analysis and allows the interpretation of the status of each variable directly.

\subsection{FDI Scheme}
\label{sec:fault-detection-and-identification-scheme}

\begin{figure*}
  \centering
  \includegraphics[width=0.78\textwidth]{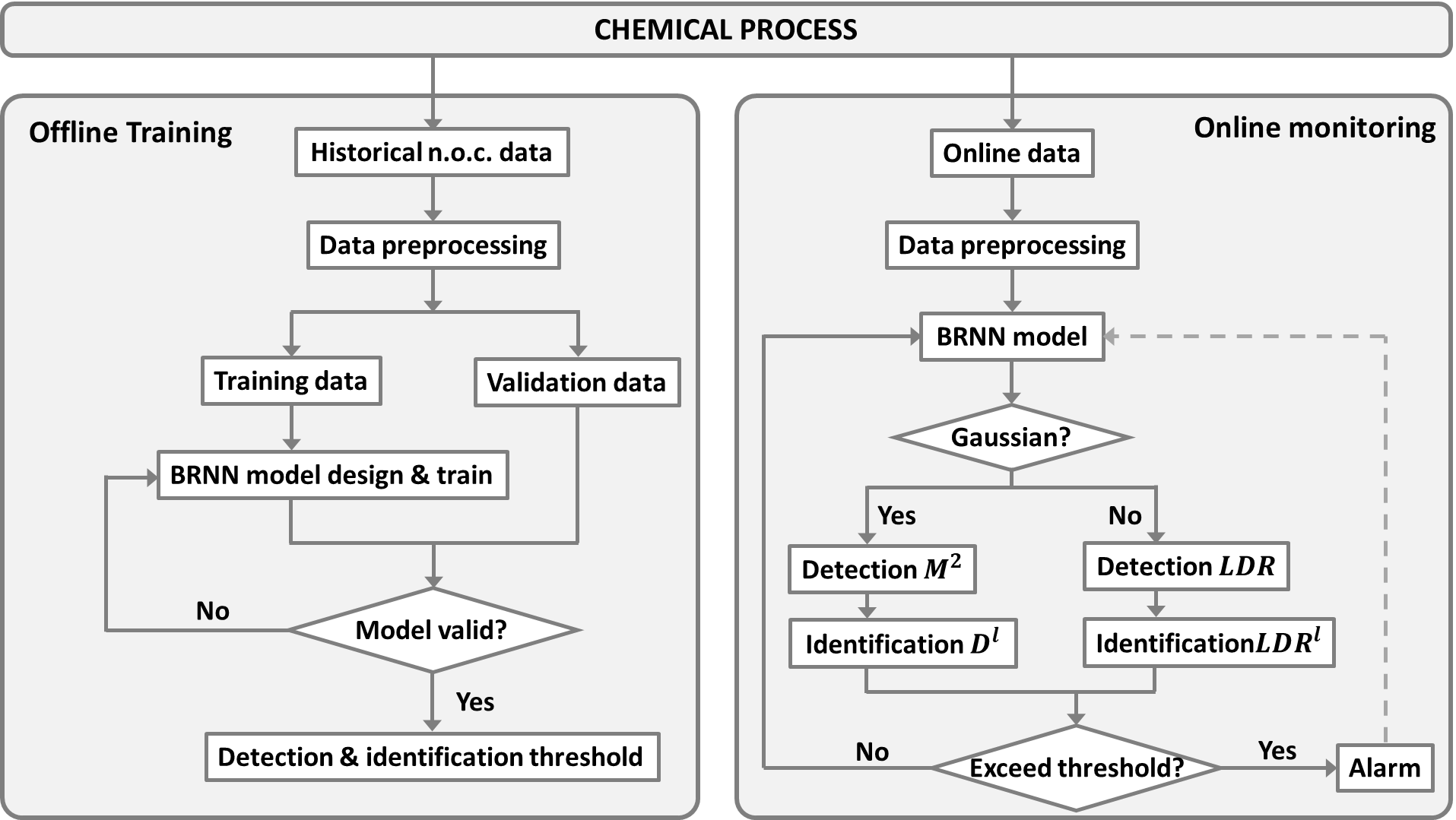}
  \caption{Flowchart of the BRNN-based FDI methodology. The offline training stage (left) and the online monitoring stage (right) are shown in the figure. The procedure starts with offline training, and then the offline-trained model is used during online monitoring. The choice of statistics for detection and identification is made at design time.}
  \label{fig:flowchart}
\end{figure*}

For completeness, the overall methodology is summarized in Figure \ref{fig:flowchart}.
Although the figure explicitly shows the two methods for detecting and identifying faults, this decision of which method to use is actually done at the design stage rather than during operations. 
In either case, the BRNN model with variational dropout is crucial to the methodology by providing samples that characterize the uncertainty and directly enable both FDI. The $M^2$ or LDR statistics are used to detect the fault in the system, while the $D^l$ or $\text{LDR}^l$ statistics are used to identify the impacted variables useful for locating the fault and possible root cause analysis. Minimal computation is needed for fault identification, having to calculate only some additional statistics on the same samples.

\section{Case Studies}\label{sec:case-studies}

In this section, the effectiveness of the proposed BRNN-based FDI method is demonstrated in two case studies: the benchmark Tennessee Eastman process synthetic dataset and a real dataset from a chemical plant. 

For comparison, results are also shown for PCA \citep{Jackson1979,Kourti1996} and DPCA \citep{Ku1995} FDI methods.
For each method, both models with and without dimension reduction are considered, and identified by prefix `r-' or `f-', respectively. For the models with reduced dimension, parallel analysis \citep{Downs1993} is used to determine the number of PCs $a$ to retain in the model.
These (D)PCA-based methods are commonly accepted benchmark methods for algorithm comparison in the FDI community \citep{Chiang2000,Yin2014,De2015, Venkatasubramanian20032}. DPCA in particular provides an interesting contrast to the proposed BRNN method because DPCA also models both spatial and temporal correlations, albeit in a limited form. As mentioned in the introduction, DPCA is limited to linear dynamics and correlations and scale poorly with increased temporal memory depth. The proposed BRNN method does not have these limitations.

In both case studies, the BRNN model constructions were implemented in TensorFlow \citep{TensorFlow}, and a number of BRNN model configurations and hyperparameters were tested. The choices included different recurrent node types (regular RNN, GRU, and LSTM cells), activation functions (i.e., linear, sigmoid, hyperbolic tangent, and rectifier linear), number of recurrent nodes/states $m_s$, number of recurrent layers, regularization hyperparameter values $\lambda$, dropout probabilities $p_{d}$, and RNN training parameters (e.g., learning rate). BRNN models were trained for each variation of these configurations and hyperparameters. The final model configuration and hyperparameters were selected as the ones that gave the maximum likelihood on the validation dataset. Only the results for the final BRNN model are shown in the next sections.

\subsection{Tennessee Eastman Process}\label{sec:tep}

\begin{figure}
  \centering
  \includegraphics[width=0.74\textwidth]{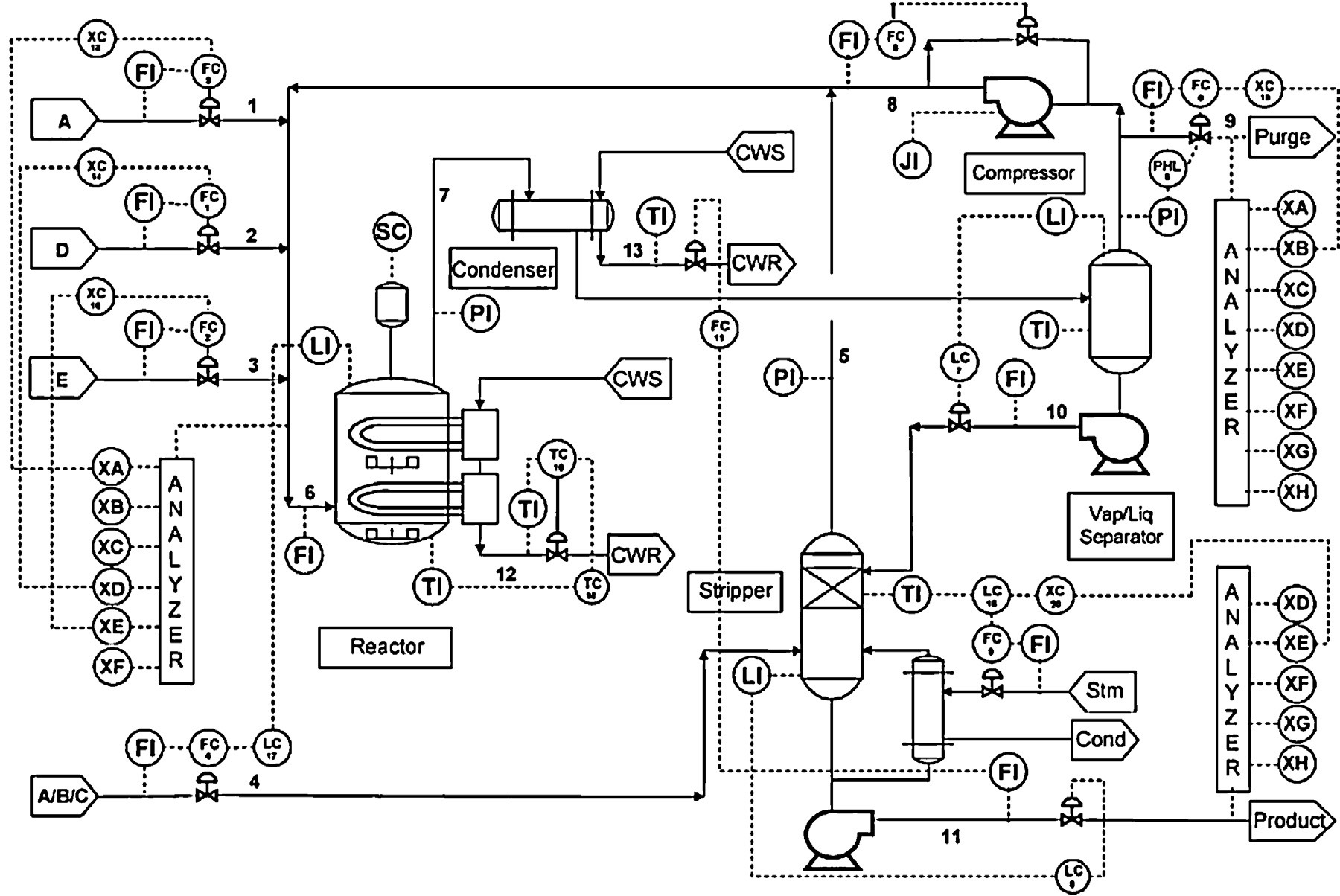}
  \caption{A process flowsheet for the TEP with the second control structure in \citep{Downs1993}.}
  \label{fig:TEP.flowsheet}
\end{figure}

The Tennessee Eastman process (TEP) is a well-known benchmark by the Eastman Chemical Company for process monitoring and control studies. It is based on a realistic industrial process with properly modified components, kinetics, and operating conditions~\citep{Downs1993}. In this study, the second plant-wide control strategy was utilized, with the process flowsheet as shown in \figref{fig:TEP.flowsheet}. The process contains eight components (A, B, C, D, E, F, G, and H) and five major units (a reactor, condenser, compressor, separator and stripper).

In this case study, $m_x = 52$ variables are used to construct the monitoring system, of which 41 are sensor measurements (XMEAS(1)--XMEAS (41)) and 11 are manipulated variables (XMV(1)--XMV(11)). During the NOC, the system is operating under one production mode and the sampling period is set to 3 min. The training data contains 480 samples and the validation data contains 960 samples. The TEP simulation contains 21 preprogrammed faults with different disturbance types and locations. Once a fault is introduced in the system, the system will either behave normally if the control system is effective in controlling the disturbance, or it will evolve outside the NOC region. For each set of data with a fault condition, the simulator first runs for 160 time points in the normal state, and then the corresponding fault disturbance is introduced with the simulator continuing to run for another 800 samples. The dataset used in this case study can be downloaded from the website of Prof.\ Richard Braatz \citep{TEPBraatz}. For further details about the TEP dataset, the reader is referred to \citet{Downs1993} or \citet{Chiang2000}.

Although a number of BRNN model architecture variations were tried as previously mentioned, the final BRNN model used in this case study contains one recurrent layer with regular RNN cell and linear activation function. A linear dense layer is used for the output layer, as is commonly done for regression tasks. This structure means that, in this case, the BRNN model implements a probabilistic linear state-space model. Although this architecture is simpler, it is also easier to train and achieved better performance than more complex structures and neuron types. The results are likely due to the fact that the inherent correlations and dynamics in the NOC data of TEP are well modeled as linear \citep{WeikeThesis}. The final model has $m_s = 80$ hidden nodes in the recurrent layer, and is trained with regularization parameter $\lambda = 10^{-4}$ and dropout rate $p_{d} = 0.1$. Given the linear structure of the model and by the central limit theorem, the predictive distribution by the final BRNN model is well approximated by the Gaussian distribution in this case. Thus, the $M^2$ and $D^l$ statistics are used for FDI. In any event, the results are quite similar for a number of configurations of the hyperparameters within a reasonable tuning range.

For the reduced dimensionality (D)PCA models used in the comparison, the number of PCs determined by parallel analysis is $a = 12$ for r-PCA and $a = 25$ for r-DPCA (with $\text{lag}=1$). The fault detection procedure for r-PCA and r-DPCA use both the $T^2$ and $Q$ statistics, whereas f-PCA and f-DPCA use only the $T^2$ statistic for fault detection, which plays the same role as the $M^2$ statistic. Contribution plots are used for (D)PCA-based fault identification. For implementation details on the (D)PCA methods, the reader is referred to \citep{Russell2000}, \citep{Zhu2014}, or \citep{Chiang2000}.

The false alarm rate~(FAR) and fault detection rate~(FDR) are used to evaluate the fault detection performance of different algorithms:
\begin{align}
  \text{FAR} &= \frac{\mbox{\# of samples with alarm during \NOC}}{\mbox{total \# of samples during \NOC}} \\
  \text{FDR} &= \frac{\mbox{\# of samples with alarm after the fault is introduced in the system}}{\mbox{total \# of samples after the fault is introduced in the system}}
\end{align}
In words, the FAR corresponds to the frequency of spurious detection of faults under NOC, and FDR is the sample frequency of a fault being detected when a fault situation is present.

The dataset contains three types of faults: controllable faults, back-to-control faults, and uncontrollable faults. Controllable faults are disturbances that can be well compensated by the control system, and therefore the disturbance does not significantly affect the process state. In these situations, since the operator is not required to intervene, the FDR should ideally be as low as the FAR to avoid distracting the operators. Back-to-control faults are disturbances that are large enough to cause the system to initially deviate from the NOC, but for which the control system is able to compensate at least some aspects of the disturbance after some time. The process measurements return to the normal region after some time, but certain manipulated or input variables remain outside the normal regime. These represent sub-optimal or off-spec conditions that ought to be handled by an operator and to be detected accordingly. Moreover, the FDI result should accurately reflect the system state, such that its evolution back to control is apparent. Finally, uncontrollable faults are faults that cannot be handled adequately by the control system and require operator intervention. For both back-to-control and uncontrollable faults, the fault detection algorithm should ideally yield high FDR to notify the operator that the system has been disturbed outside the original NOC. It is worth noting that this ``classification'' is based on prior knowledge of the faults and used here only to facilitate the interpretation of the results; it was not used anywhere in the model training.

For fault identification, the proposed BRNN-based identification plots are compared with the (D)PCA-based contribution plots, which are shown for a representative fault of each of the aforementioned types. Note that, ideally, fault identification should accurately pinpoint the variables that are affected by the fault to provide the operators with specific information for them to analyze the situation and quickly diagnose the underlying root cause. 
Accordingly, it should be verified that no variable should be identified as abnormal for controllable faults. For back-to-control faults, the abnormal variables should first be identified, and only the corresponding tuned manipulated variables should remain identified once the system is back to control. For uncontrollable faults, the identification procedure should correctly locate the abnormal variables as soon as they are outside the NOC regime.

\subsubsection{Training and validation results on the NOC data}
\label{sec:tep:noc-data}

\begin{figure*}
  \centering
  \includegraphics[width=0.96\textwidth]{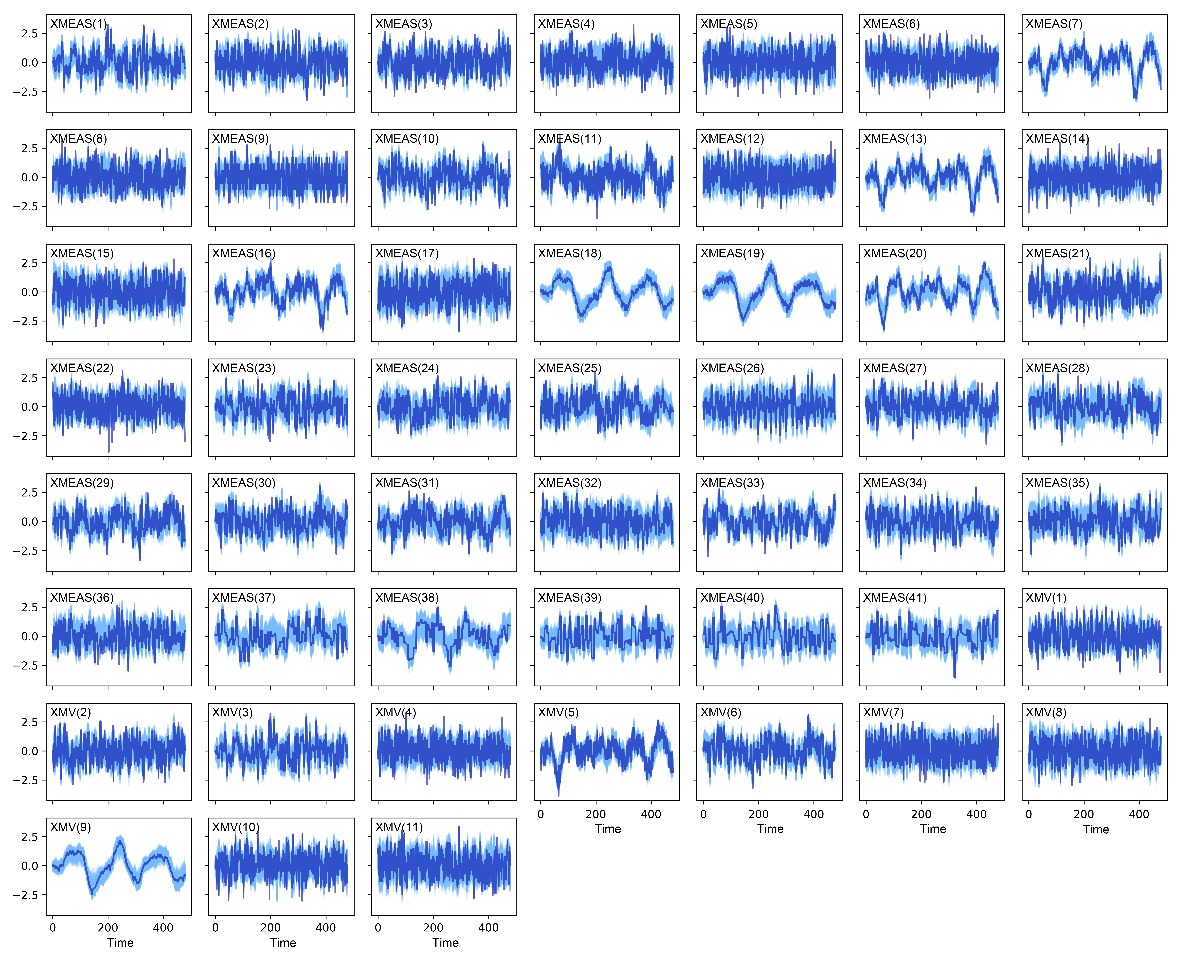}
  \caption{BRNN model outputs for TEP NOC training data. The plot shows all 52 variables in TEP. The dark blue lines are the TEP measurements and the light blue lines correspond to the BRNN model predictive distribution outputs for the NOC data. For measurements under the NOC, the dark blue lines should lie within the predictive distribution.}
  \label{fig:TEP.training}
\end{figure*}

The training and validation results are first shown to demonstrate how the posterior predictions of the BRNN model characterize the NOC.
The training results of the total 52 variables with centered and normalized values are shown in Figure \ref{fig:TEP.training}. The dark blue lines are the real data and the light blue lines are the posterior prediction samples by the BRNN model with $N = 400$ model samples by variational dropout. The results do not differ significantly for $N>100$.
When the real measurements are within the predictive distribution (in dark and light blue in the figures, respectively), the system is considered normal. This condition can be observed in Figure \ref{fig:TEP.training}, which indicates that the trained BRNN model accurately captures the NOC pattern.

\begin{figure*}
  \centering
  \includegraphics[width=0.96\textwidth]{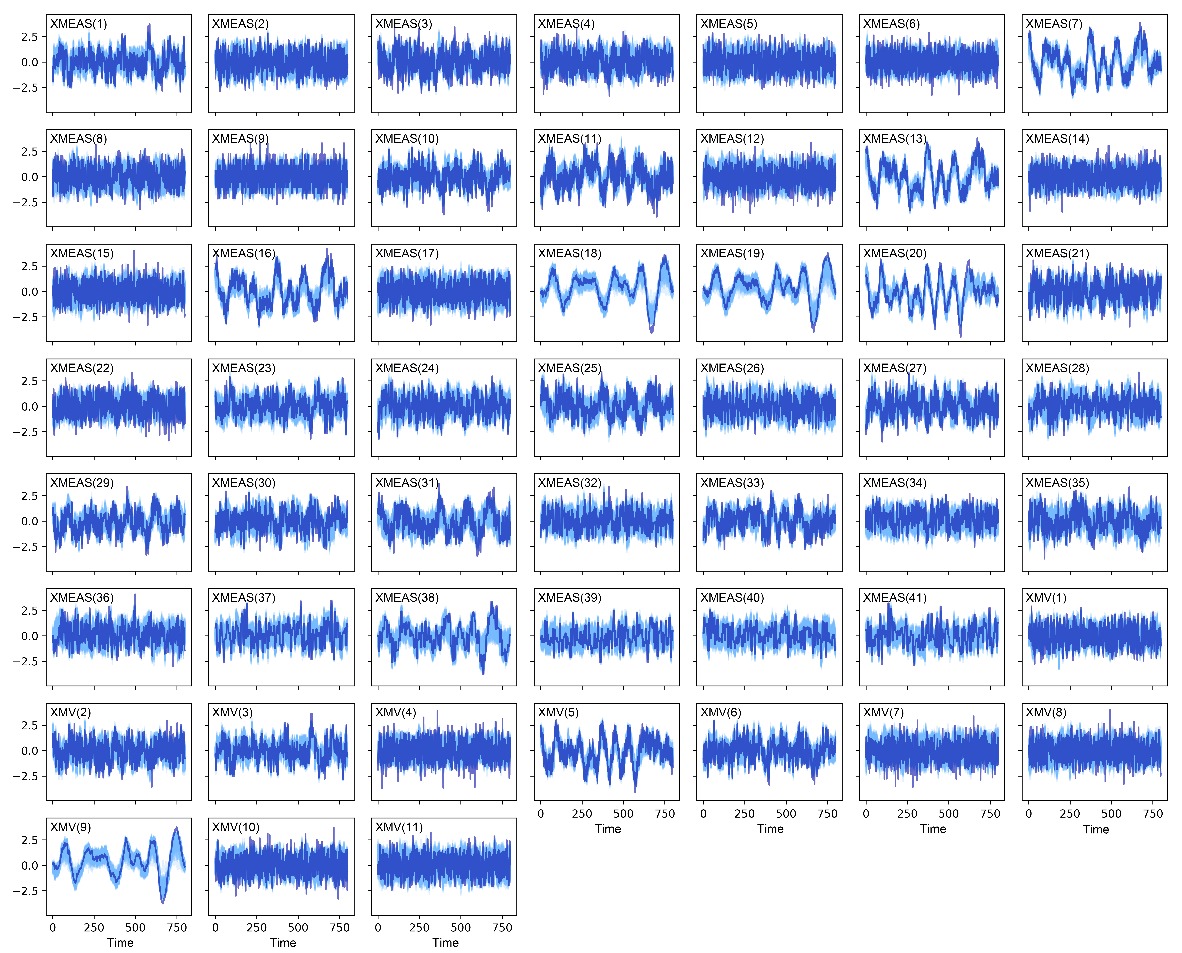}
  \caption{BRNN model outputs for TEP NOC validation data.}
  \label{fig:TEP.validation}
\end{figure*}

Then, to validate the model, the trained BRNN model is applied to a separate NOC validation dataset. These results are shown in Figure \ref{fig:TEP.validation}. As observed in the training results, the real validation measurements lie within the predictive distribution. This result indicates that the model generalizes well, meaning that it is able to capture the normal pattern without overfitting to the training data, which is crucial to avoiding high FARs.

\subsubsection{Fault detection results}\label{sec:tep:fault.detection}

\begin{table}[h]
\centering
\caption{TEP fault detection percentage results. The FAR is shown for the NOC (in the first row), and the FDR is given for the 21 fault conditions.}%
\label{tab:tep:detection.results}
\resizebox{\columnwidth}{!}{\begin{tabular}{ccccccc}
\hline
\multicolumn{1}{c}{Type}                                                        & \multicolumn{1}{c}{Fault ID} & \multicolumn{1}{c}{BRNN} & \multicolumn{1}{c}{\begin{tabular}[c]{@{}c@{}}r-PCA\\ ($a=12$)\end{tabular}} & \multicolumn{1}{c}{\begin{tabular}[c]{@{}c@{}}f-PCA\\ ($a=52$)\end{tabular}} & \multicolumn{1}{c}{\begin{tabular}[c]{@{}c@{}}r-DPCA\\ ($a=25$)\end{tabular}} & \multicolumn{1}{c}{\begin{tabular}[c]{@{}c@{}}f-DPCA\\ ($a=104$)\end{tabular}} \\ \hline
NOC                                                                       & --                             & 4.75                      & 5.00                                                                        & 4.88                                                                        & 5.00                                                                         & 5.00                                                                          \\ \hline
\multirow{3}{*}{\begin{tabular}[c]{@{}c@{}}controllable\\ faults\end{tabular}}    & IDV(3)                        & 5.00                      & 7.00                                                                        & 19.75                                                                       & 6.12                                                                         & 22.25                                                                         \\
                                                                                  & IDV(9)                        & 5.00                      & 7.88                                                                        & 15.25                                                                       & 8.87                                                                         & 21.37                                                                         \\
                                                                                  & IDV(15)                       & 7.12                      & 10.62                                                                       & 26.87                                                                       & 11.13                                                                        & 36.63                                                                         \\ \hline
\multirow{3}{*}{\begin{tabular}[c]{@{}c@{}}back to control\\ faults\end{tabular}} & IDV(4)                        & 100.00                    & 98.88                                                                       & 100.00                                                                      & 100.00                                                                       & 100.00                                                                        \\
                                                                                  & IDV(5)                        & 100.00                    & 32.62                                                                       & 100.00                                                                      & 34.50                                                                        & 100.00                                                                        \\
                                                                                  & IDV(7)                        & 100.00                    & 100.00                                                                      & 100.00                                                                      & 100.00                                                                       & 100.00                                                                        \\ \hline
\multirow{15}{*}{\begin{tabular}[c]{@{}c@{}}uncontrollable\\ faults\end{tabular}} & IDV(1)                        & 99.75                     & 99.75                                                                       & 100.00                                                                      & 99.75                                                                        & 99.25                                                                         \\
                                                                                  & IDV(2)                        & 99.00                     & 98.75                                                                       & 99.12                                                                       & 98.62                                                                        & 99.12                                                                         \\
                                                                                  & IDV(6)                        & 100.00                    & 100.00                                                                      & 100.00                                                                      & 100.00                                                                       & 100.00                                                                        \\
                                                                                  & IDV(8)                        & 98.12                     & 98.00                                                                       & 98.25                                                                       & 97.75                                                                        & 98.38                                                                         \\
                                                                                  & IDV(10)                       & 87.38                     & 54.13                                                                       & 93.50                                                                       & 55.75                                                                        & 94.63                                                                         \\
                                                                                  & IDV(11)                       & 74.75                     & 74.25                                                                       & 87.25                                                                       & 80.75                                                                        & 92.75                                                                         \\
                                                                                  & IDV(12)                       & 99.75                     & 99.00                                                                       & 100.00                                                                      & 99.25                                                                        & 100.00                                                                        \\
                                                                                  & IDV(13)                       & 95.75                     & 95.50                                                                       & 95.75                                                                       & 95.50                                                                        & 96.25                                                                         \\
                                                                                  & IDV(14)                       & 100.00                    & 100.00                                                                      & 100.00                                                                      & 100.00                                                                       & 100.00                                                                        \\
                                                                                  & IDV(16)                       & 90.38                     & 46.50                                                                       & 95.50                                                                       & 48.50                                                                        & 97.00                                                                         \\
                                                                                  & IDV(17)                       & 96.13                     & 93.13                                                                       & 97.75                                                                       & 94.78                                                                        & 98.12                                                                         \\
                                                                                  & IDV(18)                       & 90.63                     & 90.38                                                                       & 91.50                                                                       & 90.50                                                                        & 92.87                                                                         \\
                                                                                  & IDV(19)                       & 88.25                     & 25.12                                                                       & 96.00                                                                       & 34.00                                                                        & 99.50                                                                         \\
                                                                                  & IDV(20)                       & 78.63                     & 58.25                                                                       & 92.13                                                                       & 61.75                                                                        & 92.37                                                                         \\
                                                                                  & IDV(21)                       & 48.00                     & 48.50                                                                       & 61.62                                                                       & 47.88                                                                        & 59.38      \\ \hline                                                                  
\end{tabular}}
\end{table}

The fault detection results for the 21 predefined faults are shown in Table \ref{tab:tep:detection.results}. The results are grouped according to one of the above-mentioned three types of faults. For all algorithms, the FDRs are estimated with regard to the threshold estimated for a FAR of 5\%, and validated on NOC data as shown on the first row of the table.

As shown in Table \ref{tab:tep:detection.results}, the proposed BRNN-based method yields close to 5\% FDR on controllable faults, which is almost as low as the pre-determined FAR level. On the other hand, (D)PCA-based methods are overly sensitive in these cases, especially the models without model reduction, f-PCA and f-DPCA. These results show that (D)PCA-based methods cannot accurately differentiate the controllable faults from the other cases, because they do not appropriately characterize the dynamics of NOC, such as to determine if the situation is ultimately controllable.
As previously explained, controllable faults should not trigger an alert because they are handled directly by the control system. The ability of the fault detection approach to differentiate between these situations is of crucial practical importance because alerts due to these situations will often be perceived as false alarms and can erode an operator's confidence in the method and the significance of its alerts. The BRNN method is observed to be more robust to controllable fault than the (D)PCA methods.

For both back-to-control and uncontrollable faults, the BRNN method reliably detected faults with high FDRs.
Full PCA and DPCA models with the squared Mahalanobis distance were also able to detect the back-to-control and uncontrollable faults with high FDR. However, the (D)PCA models were overly sensitive for general fault detection purposes because they overreacted to controllable faults. Compared to the BRNN method, (D)PCA models emphasized higher sensitivity to disturbances at the expense of an increased likelihood of unwarranted alerts.
The reduced dimensionality (D)PCA models (i.e., r-PCA and r-DPCA), with the number of PCs determined by parallel analysis, responded more reasonably to controllable faults but also yield much worse performance compared to the BRNN method. In fact, they fail to reliably detect several back-to-control and uncontrollable faults (Faults 5, 16, and 19, for example).

It is insightful to consider how the temporal dynamics interact with the detection approach to lead to the measured FDR results. 
If a disturbance is such that the measurements oscillate around the NOC region, there will be moments in time that are momentarily indistinguishable from those in the NOC region. Since the BRNN is trained such that its state characterizes the NOC distribution in state space, it is understandable that some of these time points may not be detected as faulty. This observation explains the slightly lower FDR of the BRNN method for those cases.
Unlike the BRNN, the (D)PCA methods do not model internal system dynamics under the NOC. Hence, since the internal dynamics are not considered when explaining the observed data, these models do not have this detection ambiguity. This lack of ambiguity is achieved at the expense of the inability by (D)PCA to assess whether a fault is controllable.
In summary, the fault detection results indicate the BRNN has high detection accuracy and is able to more robustly detect faults when operator intervention is truly necessary.

Specific FDI results are presented and discussed in detail below. Faults~1, 3, and 5 are presented because they are representative of each type. There was no significant differences between fault within the same type. The use of the BRNN contribution plot results for fault propagation analysis and one example is given for Fault 6.

\subsubsection{Controllable fault: Fault 3}%
\label{sec:tep:fault3}

Fault 3 is considered first as a demonstrative controllable fault.
For Fault 3, the D feed temperature in Stream 2 has a step change at the $160^{\mathrm{th}}$ time point. Since this change in feed temperature is handled immediately and directly by the control system, the process is not driven outside its normal operating state. In this case, a data-driven fault detection algorithm should not trigger the alarm (beyond the chosen FAR).

\begin{figure*}
  \centering
  \includegraphics[width=0.96\textwidth]{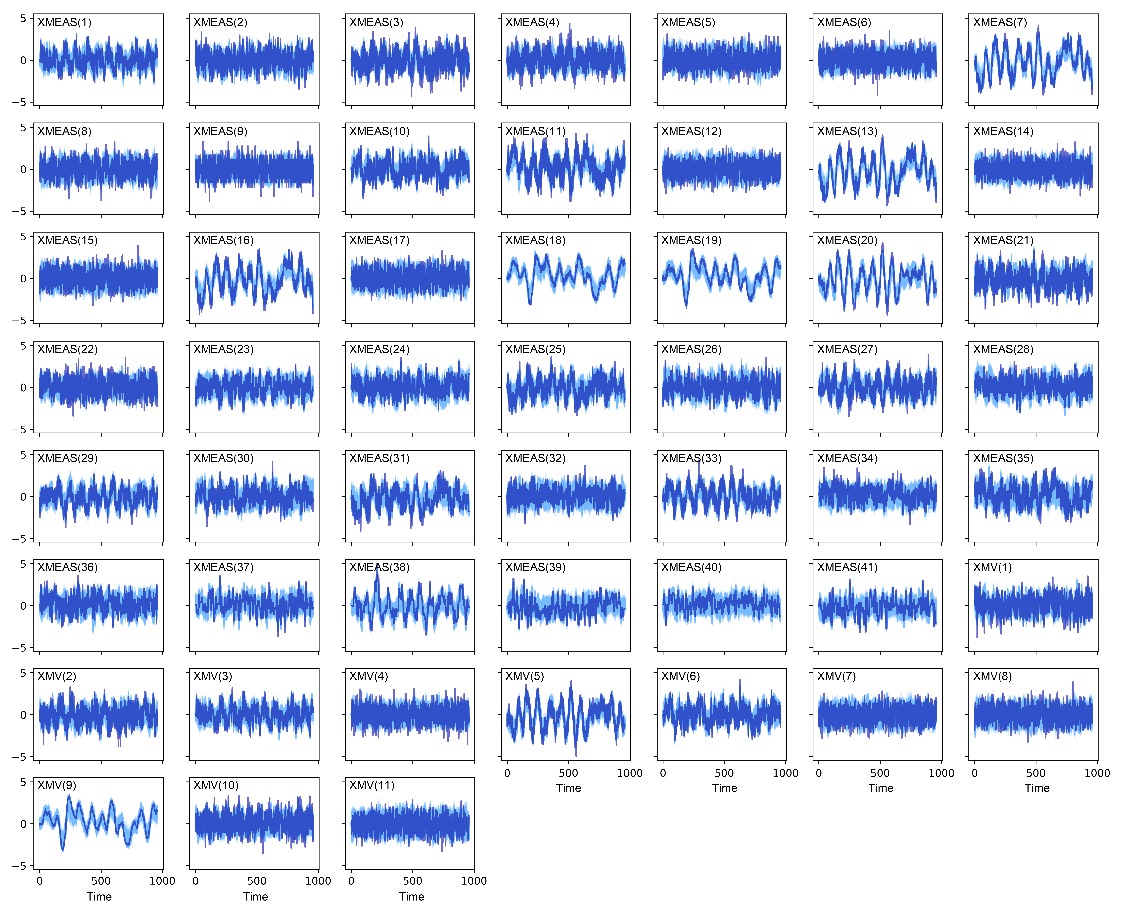}
  \caption{BRNN model outputs for TEP Fault 3.}
  \label{fig:TEP:fault3.outputs}
\end{figure*}

The prediction results by the BRNN model are shown in Figure \ref{fig:TEP:fault3.outputs}. Similarly to the NOC case, the dark blue lines (i.e., real measurements) are within the distribution high-likelihood area characterized by the light blue lines, which indicates that the system is operating under the NOC.

\begin{figure}
  \centering
  \includegraphics[width=0.48\textwidth]{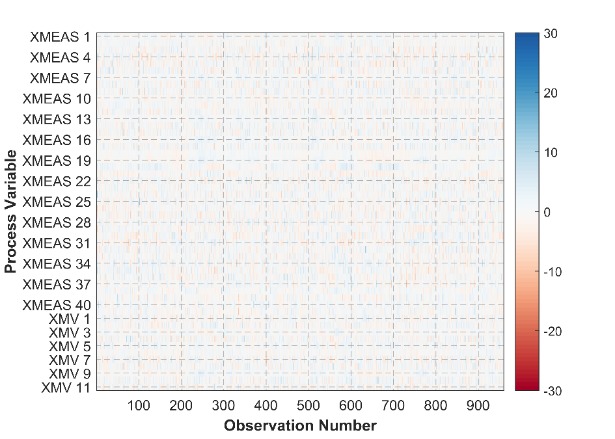}
  \caption{Fault identification plot of the BRNN-$D^l$ statistic for Fault 3. The $\left\{ D^l \right\}_{l = 1,\dots,52}$ values for the 960 timesteps are color coded in the identification plot. Variables with dark blues have high values of $D^l$, meaning that the variable has positively deviated from the NOC region. Conversely, variables with dark red have low values of $D^l$ and have negatively deviated from the NOC region. A light color means the variable is not significantly affected by the disturbance. As expected, no variable significantly deviates from the NOC.}
  \label{fig:TEP:fault3.idplot}
\end{figure}

\begin{figure*}
  \centering
  \includegraphics[width=0.96\textwidth]{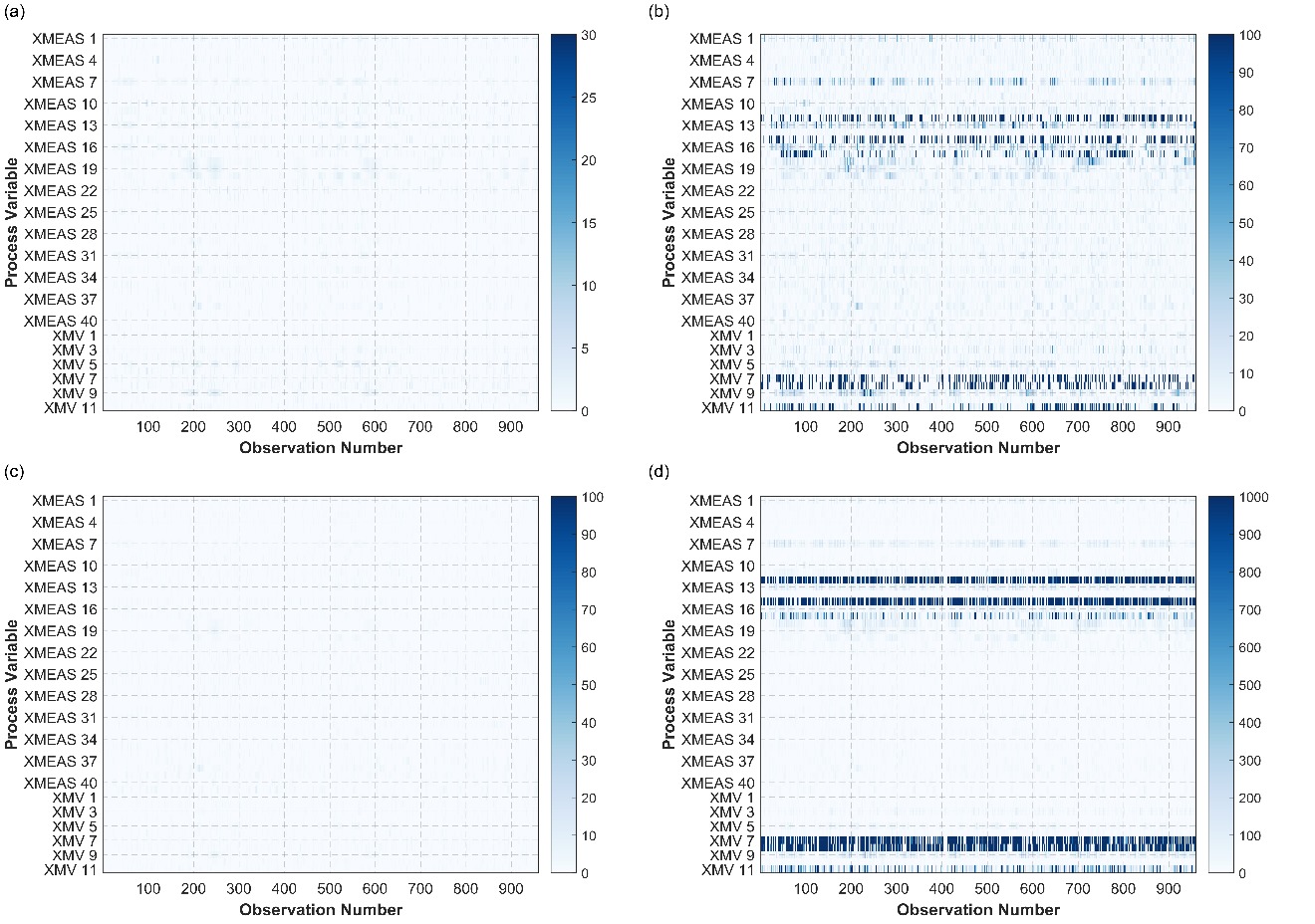}
  \caption{Contribution plots for Fault 3 from (a) r-PCA, (b) f-PCA, (c) r-DPCA, and (d) f-DPCA. The plot shows the contribution factor and with the darkness of the blue color indicating the amount of deviation of the variable from the NOC region.}
  \label{fig:TEP:fault3.contribplot}
\end{figure*}

Fault identification results by the BRNN and (D)PCA methods are shown in Figures \ref{fig:TEP:fault3.idplot} and \ref{fig:TEP:fault3.contribplot}, respectively. The color indicates the deviation from the NOC over time for each of the 52 variables. The $D^l$ statistic (c.f.\ Equation \ref{eq:fault.id:Dl.stat}) is used in the BRNN identification plot.
As shown in Figure \ref{fig:TEP:fault3.idplot}, the BRNN model identifies that no variable has its normal operating dynamics significantly affected by Fault 3, as is expected. In contrast, the contribution plots in Figure \ref{fig:TEP:fault3.contribplot}bd, by the full PCA and DPCA models, incorrectly identify several variables as being affected by the disturbance even before the introduction of the disturbance (at the $160^{\mathrm{th}}$ sample). This further demonstrates the oversensitiveness of those models. The r-PCA and r-DPCA models have identification results that are similar to those of the BRNN model and are somewhat robust to controllable faults, but at the expense of robustness in fault detection.

In summary, for controllable faults, BRNN-based FDI is robust and successfully characterizes those disturbances as corresponding to NOC. r-PCA and r-DPCA methods gave similar fault identification results but have lower detection rates (c.f.\ Table \ref{tab:tep:detection.results}). The f-PCA and f-DPCA models were clearly oversensitive, have high FARs, and incorrectly characterized the controllable faults in the contribution plots.

\subsubsection{Back-to-control fault: Fault 5}%
\label{sec:tep:fault5}

Fault 5 is a representative example of a back-to-control fault.
This fault involves a step change in condenser cooling water inlet temperature. This step change requires a step change in the condenser cooling water flow rate XMV(11) by the control system. While the fault is ultimately controllable, the fault causes the system at first to operate off-spec, or at least sub-optimally. In this particular, immediately after the fault occurs, the system oscillates with about 32 variables exhibiting this similar transient oscillation behavior. The process returns to control after about 10~hours, at which point the sensor measurements XMEAS(1)--XMEAS(41) are back to their pre-disturbance set-points, and only the manipulated variable XMV(11) remains outside the NOC regime, tuned so as to compensate the step change in condenser cooling water inlet temperature. 

\begin{figure*}
  \centering
  \includegraphics[width=0.96\textwidth]{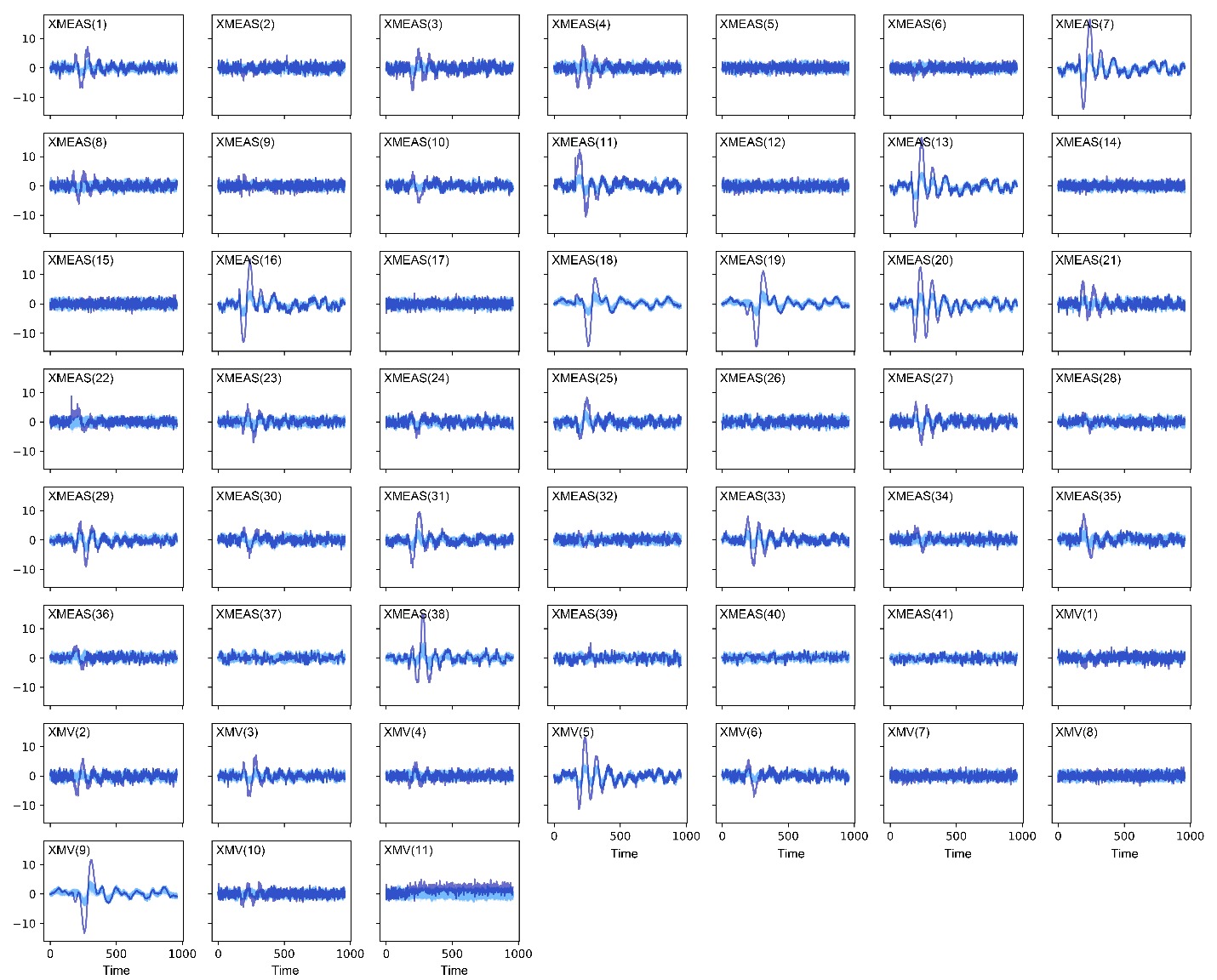}
  \caption{BRNN model outputs for TEP Fault 5.}
  \label{fig:TEP:fault5.outputs}
\end{figure*}

The BRNN results for all of the variables are shown in Figure \ref{fig:TEP:fault5.outputs}. As expected, when the fault is introduced, several measurements in dark blue lines deviate from the posterior predictive distribution under the NOC shown in light blue. After about 200 data points the system returns to control, verified by the fact that all the system measurements (XMEAS(1)--XMEAS(41)) are back within the predictive NOC region while only the manipulated variable XMV(11) maintains a systematic deviation off-the-center of the BRNN model prediction distribution.
These results show that the BRNN model is able to correctly identify the NOC pattern and how the deviation from the predictive distribution accurately locates the faulty variable under disturbance.
The BRNN model is also able to better assess the state of the system, distinguishing the back to control faults from the uncontrollable faults by showing the transient deviation of the process variables and their return to the NOC region.

\begin{figure}
  \centering
  \includegraphics[width=0.48\textwidth]{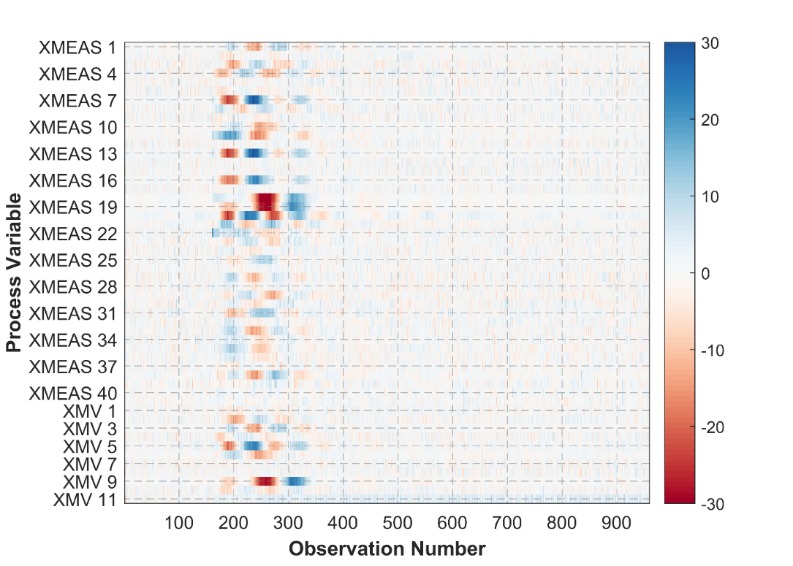}
  \caption{Fault identification plot by BRNN-$D^l$ for Fault 5. The
    switch between dark blue and red colors shows that the system is
    undergoing large fluctuation.}
  \label{fig:TEP:fault5.idplot}
\end{figure}

The fault identification plot by BRNN is shown in Figure \ref{fig:TEP:fault5.idplot}. This example showcases the typical pattern of back to control faults, with several measurements outside the predictive region after the fault is introduced and only the manipulated variables deviating once the system is back to steady state. In this case, about 32 variables are affected once the disturbance is introduced to the system, and the color switches between blue and red, indicating the system is oscillating. The plot also clearly shows how, after the $360^{\mathrm{th}}$ time point, all system variables except XMV(11) are undoubtedly back to normal. XMV(11) remains consistently above the predictive mean after the fault as that is forced by the controller to compensate for the fault. However, the magnitude of the deviation of XMV(11) is relatively small, indicating that the disturbance is no longer critical.

The BRNN identification plot also contains crucial information for locating the likely root cause of the fault. The plot clearly shows that XMEAS(22) is the first variable positively deviated from the predictive distribution, which indicates the higher than normal separator cooling water outlet temperature. Combined with the fact that the condenser cooling water flow rate is increased to compensate for the disturbance to the system, one would reason that the root cause is the increase in the condenser cooling water temperature. After the condenser cooling water temperature increases, the outlet stream from the condenser to the separator also increases the temperature, resulting in an increase in the temperature in the separator, which finally results in the increase in separator cooling water outlet temperature.

\begin{figure*}
  \centering
  \includegraphics[width=0.96\textwidth]{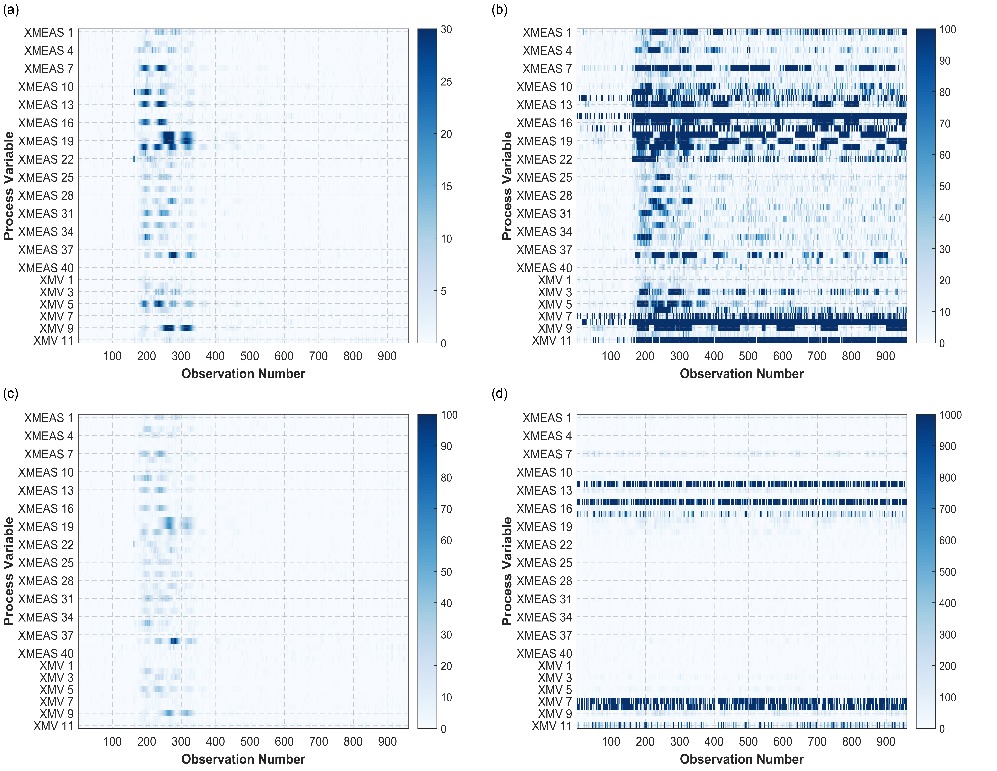}
  \caption{Contribution plots for Fault 5 from (a) r-PCA, (b) f-PCA, (c) r-DPCA, and (d) f-DPCA.}
  \label{fig:TEP:fault5.contribplot}
\end{figure*}

For comparison purposes, the contribution plots by PCA and DPCA methods are shown in Figure \ref{fig:TEP:fault5.contribplot}. The r-PCA and r-DPCA model results in Figures \ref{fig:TEP:fault5.contribplot}ac fail to identify the consistent deviation in XMV(11), which clearly explains their detection results for this fault. These results demonstrate again that the r-PCA and r-DPCA models can exhibit much lower detection and identification sensitivity than the BRNN method. The results of the f-PCA and f-DPCA models in Figures \ref{fig:TEP:fault5.contribplot}bd clearly identify the deviations in several variables. However,the  f-PCA and f-DPCA are oversensitive and identify variables in an unspecific manner, which prevents those statistics from being used, at least directly, by operators for diagnosing the root cause of the fault.

For the back-to-control fault, BRNN FDI had high accuracy and robustness. Moreover, this method yielded more specific information for evaluating the state of the system. By inspecting the identification plot, operators have a clear view about which variables are affected by the disturbance and are able to assess the type of the fault occurring and the current stage of the system.

\subsubsection{Uncontrollable fault: Fault~1}%
\label{sec:tep:fault1}

An uncontrollable fault is now considered.
Fault 1 involves a step change in the A/C feed ratio in Stream 4, which results in an increase in the C feed and a decrease in the A feed. This subsequently leads to a decrease in feed A in the recycle Stream 5 and the controller reacts by increasing the A feed flow in Stream 1. These two effects conflict with each other, thereby shifting the system to an uncontrollable operating situation.

\begin{figure*}
  \centering
  \includegraphics[width=15cm]{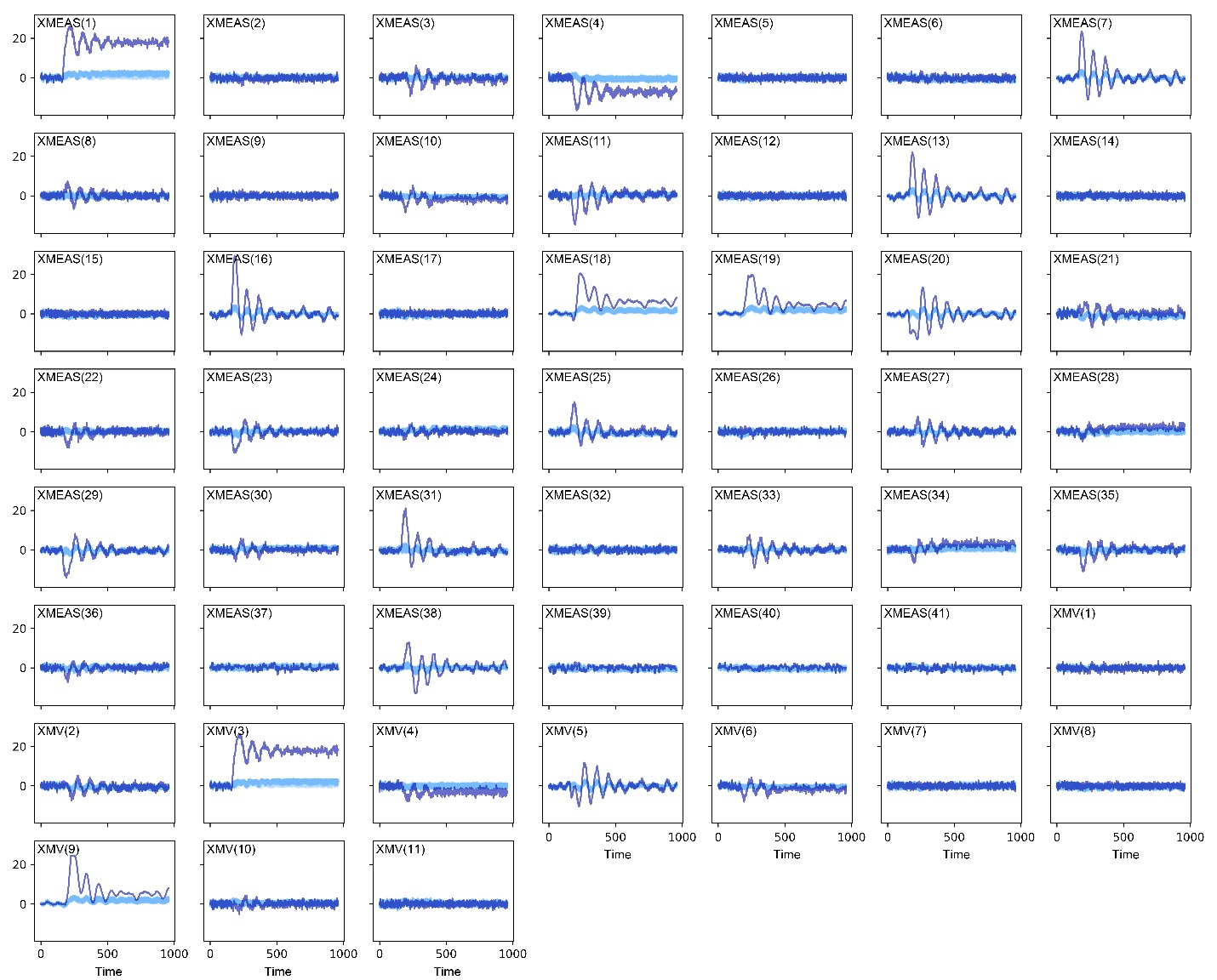}
  \caption{BRNN model outputs for TEP Fault 1.}
  \label{fig:TEP:fault1.outputs}
\end{figure*}

The BRNN model output results are shown in Figure \ref{fig:TEP:fault1.outputs}. After the fault is introduced to the system, more than half of the variables are observed to deviate significantly from the BRNN predictive NOC region. All of the (D)PCA methods are also capable of detecting this fault.

\begin{figure}
  \centering
  \includegraphics[width=0.48\textwidth]{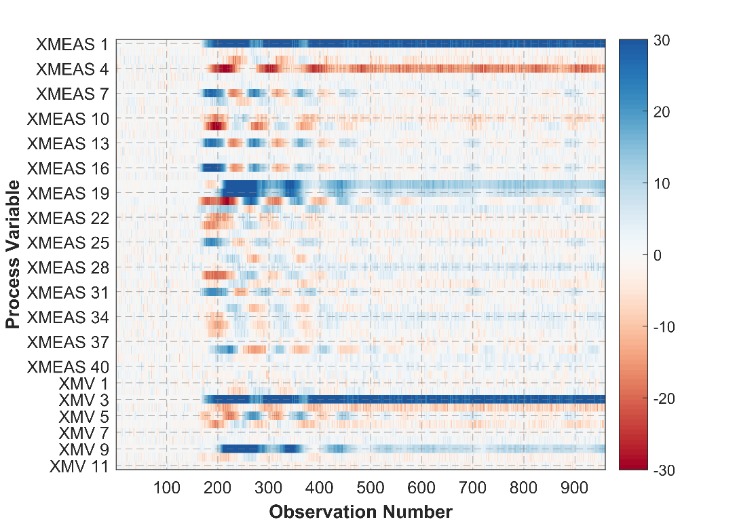}
  \caption{Fault identification plot by BRNN-$D^l$ for Fault 1.
    The root cause for this uncontrollable fault can be assessed by looking
    at the variables that are persistently off the NOC region.}
  \label{fig:TEP:fault1.idplot}
\end{figure}

The corresponding BRNN fault identification statistics are shown in Figure \ref{fig:TEP:fault1.idplot}. Since the system is seriously affected by the disturbance and several variables associated with material balances (e.g., composition, pressure) change significantly, this fault is easily detected. The long-term and uncontrollable nature of the fault on these measurements and manipulated variables can also be observed in the identification plot, making the fault easy to diagnose based on those variables.

\begin{figure*}
  \centering
  \includegraphics[width=0.96\textwidth]{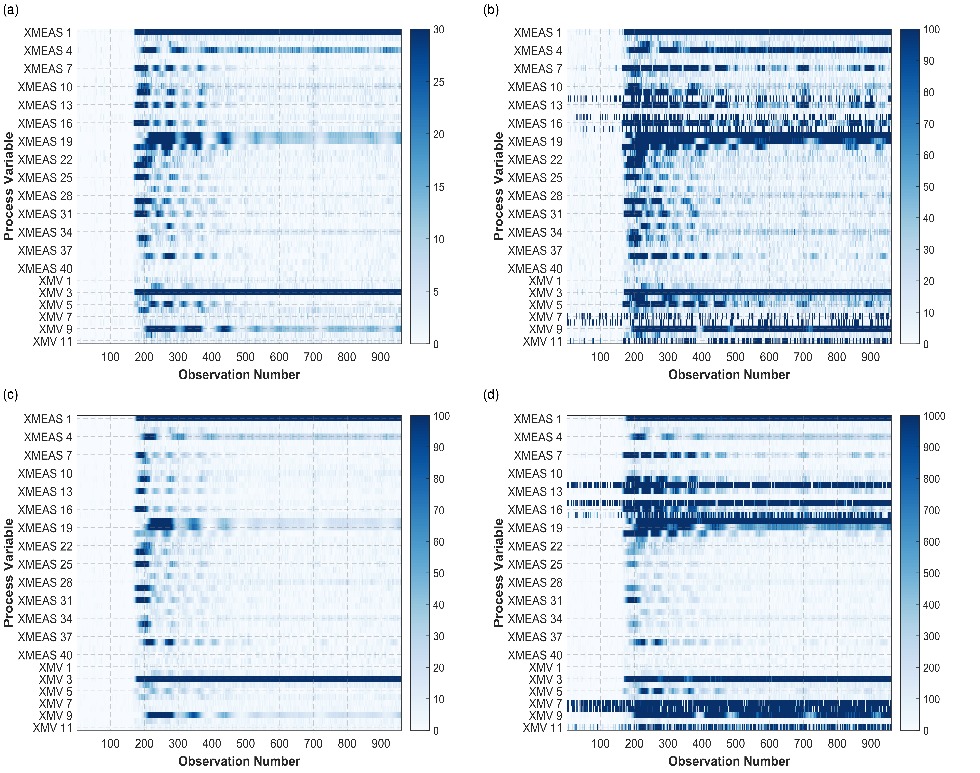}
  \caption{Contribution plots for Fault~1 from (a) r-PCA, (b) f-PCA, (c) r-DPCA, and (d) f-DPCA.}
  \label{fig:TEP:fault1.contribplot}
\end{figure*}

As before, the contribution plots by (D)PCA methods are shown in Figure \ref{fig:TEP:fault1.contribplot}. The r-PCA and r-DPCA models, in Figure \ref{fig:TEP:fault1.contribplot}ac, both give somewhat results similar to those of the BRNN model in Figure \ref{fig:TEP:fault1.idplot}. However, both of them fail to identify the continued deviation in XMV(4) (shown between XMV(3) and XMV(5) in Figure \ref{fig:TEP:fault1.contribplot}) for instance, which is the manipulated variable for total feed flow in Stream 4 and clearly plays a central role in the fault. In contrast, it is clearly identifiable from the BRNN results in Figure \ref{fig:TEP:fault1.idplot} that XMV(4) has negatively deviated from the NOC region. 
The contribution plots of f-PCA and f-PCA in Figure \ref{fig:TEP:fault1.contribplot}bd show the identification of the involved variables, but again highlight several other variables that are unrelated to the fault and operating within their normal pattern (such as XMV(11)). As previously observed, this again shows that the f-PCA and f-DPCA models are overly sensitive and their identification results require substantial additional processing such that operators cannot directly use them to diagnose the fault.

In summary, the BRNN model is able to accurately and robustly detect and identify uncontrollable faults. Perhaps most crucially, BRNN identification plots provide clear information that is directly useful for root cause analysis. While several (D)PCA models are also able to detect uncontrollable faults, their identification results are less accurate and precise than for the BRNN model.

\subsubsection{Fault propagation path analysis: Fault 6}%
\label{sec:tep:fault-propagation-fault6}

This section shows how the accuracy and specificity of the BRNN identification statistics can be used for fault propagation path analysis. The key observation is that the chronological sequence of events of when each variable deviates significantly from its NOC is useful information to understand the start and evolution of the disturbance through the process \citep{Chiang2003}. The BRNN method can extract this information with a high degree of temporal precision. This information can then be combined with expert knowledge of the process to examine the propagation of the fault through the system.

This approach is exemplified here using Fault 6, which is an uncontrollable fault induced by a loss of feed A in Stream 1. The loss of component A thus causes the control system to increase the manipulated variable XMV(3) in order to increase A in the system and attempt to compensate for the disturbance. However, since there is no component A in Stream 1, the control system fails to take the system back to NOC. Due to the severity of this fault, a large portion of system variables is affected.

\begin{figure}
  \centering
  \includegraphics[width=0.48\textwidth]{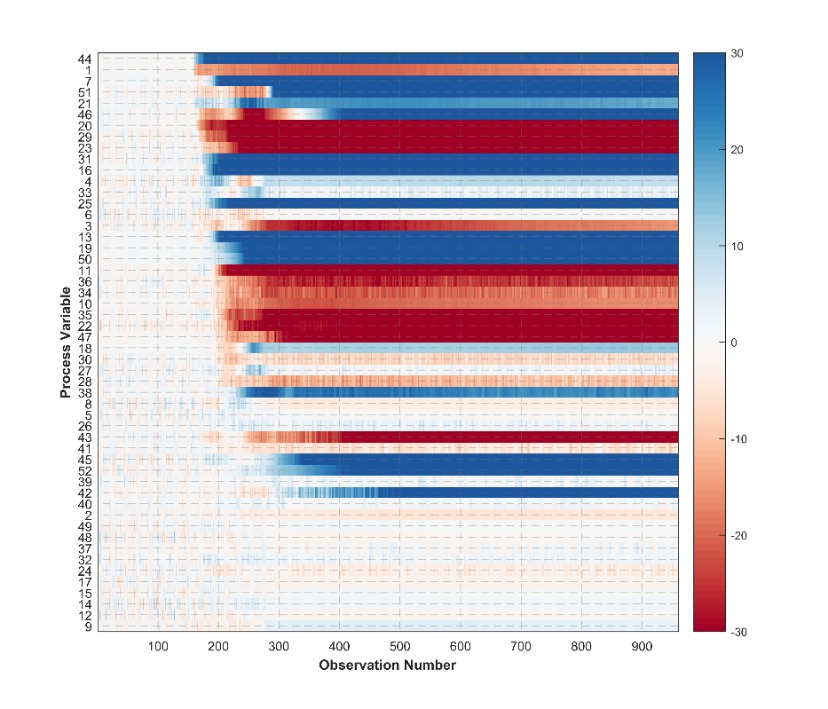}
  \caption{Sorted fault identification plot according to the detected deviation occurrence time of BRNN-$D^l$ for Fault 6.}
  \label{fig:TEP:fault6.idplot}
\end{figure}

The temporal sequence of the fault through the process is achieved by sorting the identification plot according to the time when each variable significantly deviates from the NOC. For this approach, one needs to estimate the threshold used to determine when the deviation is significant. In our case, this is estimated using the NOC validation set and determined to be $D_{\text{th}}^l = 4.8$. Then, once a fault is detected, the process variables are sorted according to the time index at which its $D^l$ statistic first exceeds the threshold, yielding the sorted identification plot shown in Figure \ref{fig:TEP:fault6.idplot}. The $y$-axis numbers 1--52 correspond to $[\text{XMEAS}(1), \dots, \text{XMEAS}(41), \text{XMV}(1), \dots, \text{XMV}(11)]$.

\begin{figure}
  \centering
  \includegraphics[width=0.74\textwidth]{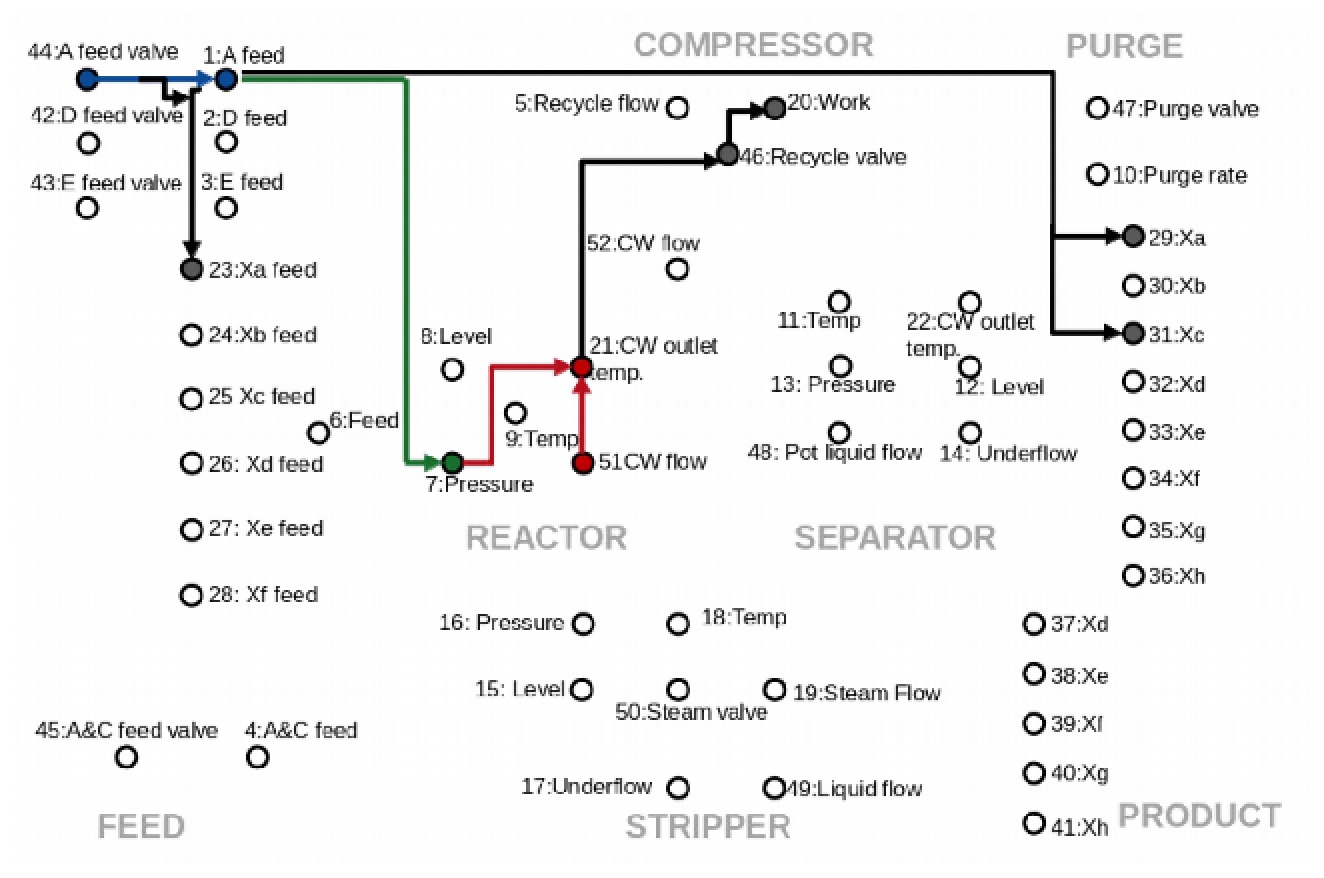}
  \caption{Fault~6 propagation path at the $180^{\mathrm{th}}$ data point (1 hour after the fault occurs). Colored nodes indicate that the corresponding variable has been detected as deviating significantly from the NOC.}
  \label{fig:TEP:fault6.propagpath}
\end{figure}

A diagram of the fault propagation path is then obtained by combining the timing results of the sorted BRNN identification plot with the knowledge of the process, as demonstrated in Figure \ref{fig:TEP:fault6.propagpath} for Fault 6. When the fault is introduced in the system, XMEAS(1) and XMV(3) are affected and deviates from the NOC first. Then, after a few minutes, the reactor pressure measurement XMEAS(7) is affected. Then the reactor cooling water system is also affected due to the change in the mass inside the reactor and both XMEAS(21) and XMV(10) deviate from the NOC. The diagram in Figure \ref{fig:TEP:fault6.propagpath} highlights that, after 1 hour, the fault has already propagated to the final product and the concentration of A and C have been affected, thus clearly showing the impact of the fault in the system at that point in time.

The approach outlined here shows how the properties of the BRNN method can be used to easily determine and visualize the fault propagation path. This information is crucial to operators to accurately diagnose the fault and determine which parts of the process have been affected.

\subsection{Real Industrial Dataset}%
\label{sec:real-data}

The next case study further demonstrates the efficiency of the proposed BRNN method on a real dataset from a chemical manufacturing process. The use of this method for real-time FDI is a promising application for the next generation of process monitoring systems in chemical plants. The complex nature of real chemical manufacturing processes and their intricate control system dynamics, make BRNN the best-suited tool to extract and recognize these patterns from data in comparison to traditional methods.

The dataset pertains to the operation of an amine tower. The column experienced foaming issues resulting in faults that decrease the efficiency of the process. There are a total of 20 sensor measurements with a sampling time of $t = 1$ min. A total of two months of data are available. Two events has been recorded by operators as a result of the foaming issue in the tower. However, it is also possible that additional disturbances are encountered during the two-month operating window that have been previously missed.

The final BRNN model uses standard RNN cells with the sigmoid activation function. There is one hidden recurrent layer with 40 units (i.e., `state variables'). The dropout probability is set to $p_d = 0.1$ and the regularization parameter to $\lambda = 10^{-5}$. The BRNN model using LSTM or GRU cells yield similar performance in spite of the higher complexity and thus those results are omitted. Similar to the TEP data, the results are similar for a number of configurations of the hyperparameters within a reasonable tuning range.

For comparison, the PCA and DPCA models, with the number of PCs determined by parallel analysis and full models without dimension reduction, are also applied. The number of PCs from parallel analysis is determined to be $a=6$ for r-PCA and $a=9$ for r-DPCA.

Due to the sensitivity of the data, the actual measurement values and the BRNN model predictions are omitted and only the detection and identification results are shown. The posterior predictive distribution is observed to be multi-modal and thus the LDR statistics are used for FDI. The number of $k$NN is set to the range of 10 to 20. The dataset is divided into a 35-day training dataset, a 17-day validation dataset, and two testing datasets. The first testing dataset spanned 7 days containing Fault 1, and the second testing dataset spanned 14 days containing Fault 2.

\begin{figure}
  \centering
  \includegraphics[width=0.48\textwidth]{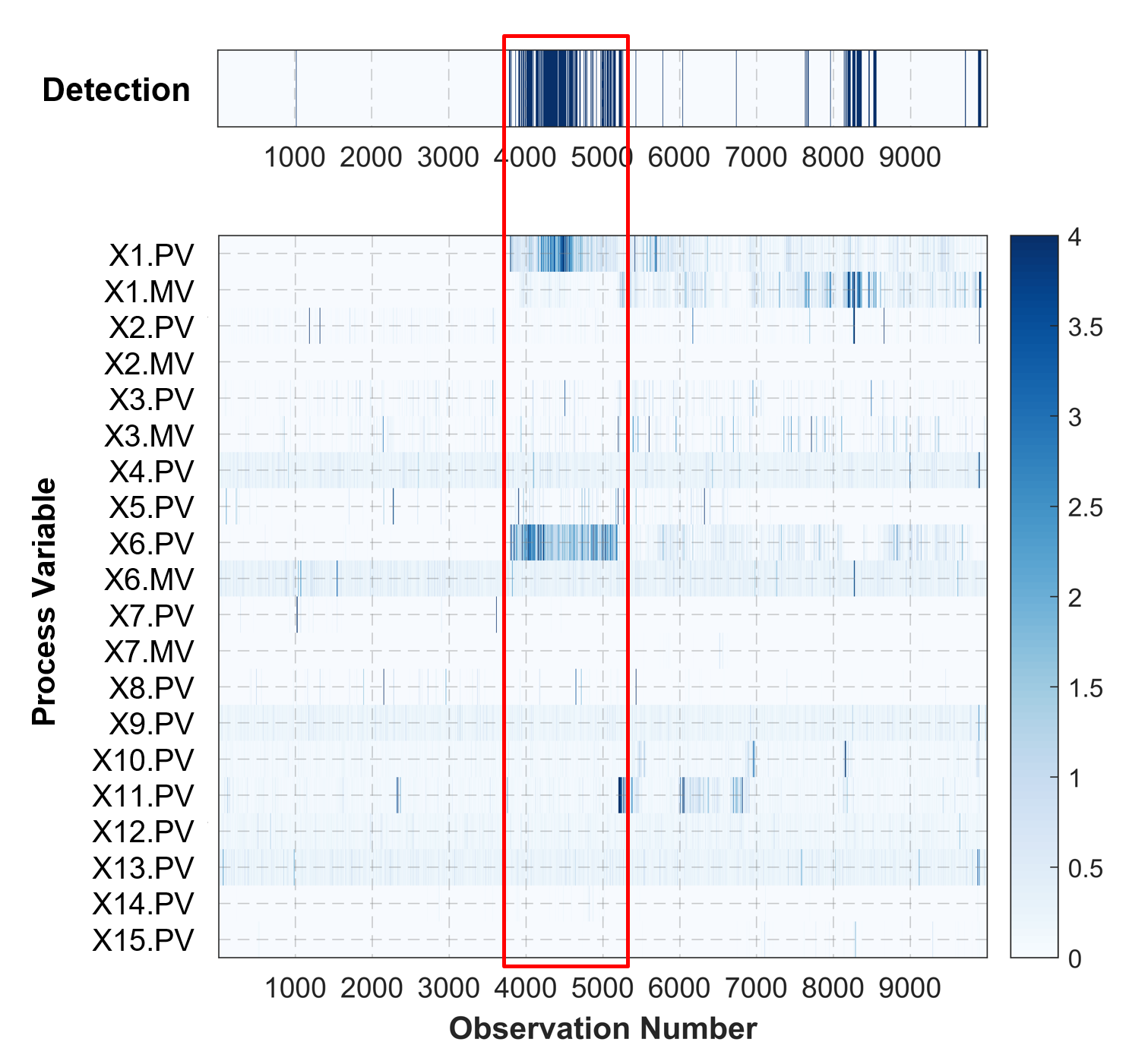}
  \caption{FDI by the BRNN model on the Fault 1 testing data. The red box indicates the period with the foaming event recorded by the operator.}
  \label{fig:fault1:brnn}
\end{figure}

\subsubsection{Fault 1 results}%
\label{sec:real-data:fault1}

The BRNN FDI results for Fault 1 are shown in Figure \ref{fig:fault1:brnn}. The BRNN model successfully detects the documented event, marked by the red box in the figure. The model also accurately pinpoints the variables that are most affected by the foaming issue, X1.PV and X6.PV. Moreover, it also highlights several points after the $8000^{\mathrm{th}}$ time point that may have been originally missed. During these later periods, the X1.MV sensor measurement is identified by the BRNN method. This is subsequently verified to have been the result of large unexplained fluctuations in that variable and that the BRNN has performed as expected.

\begin{figure}
  \centering
  \includegraphics[width=0.96\textwidth]{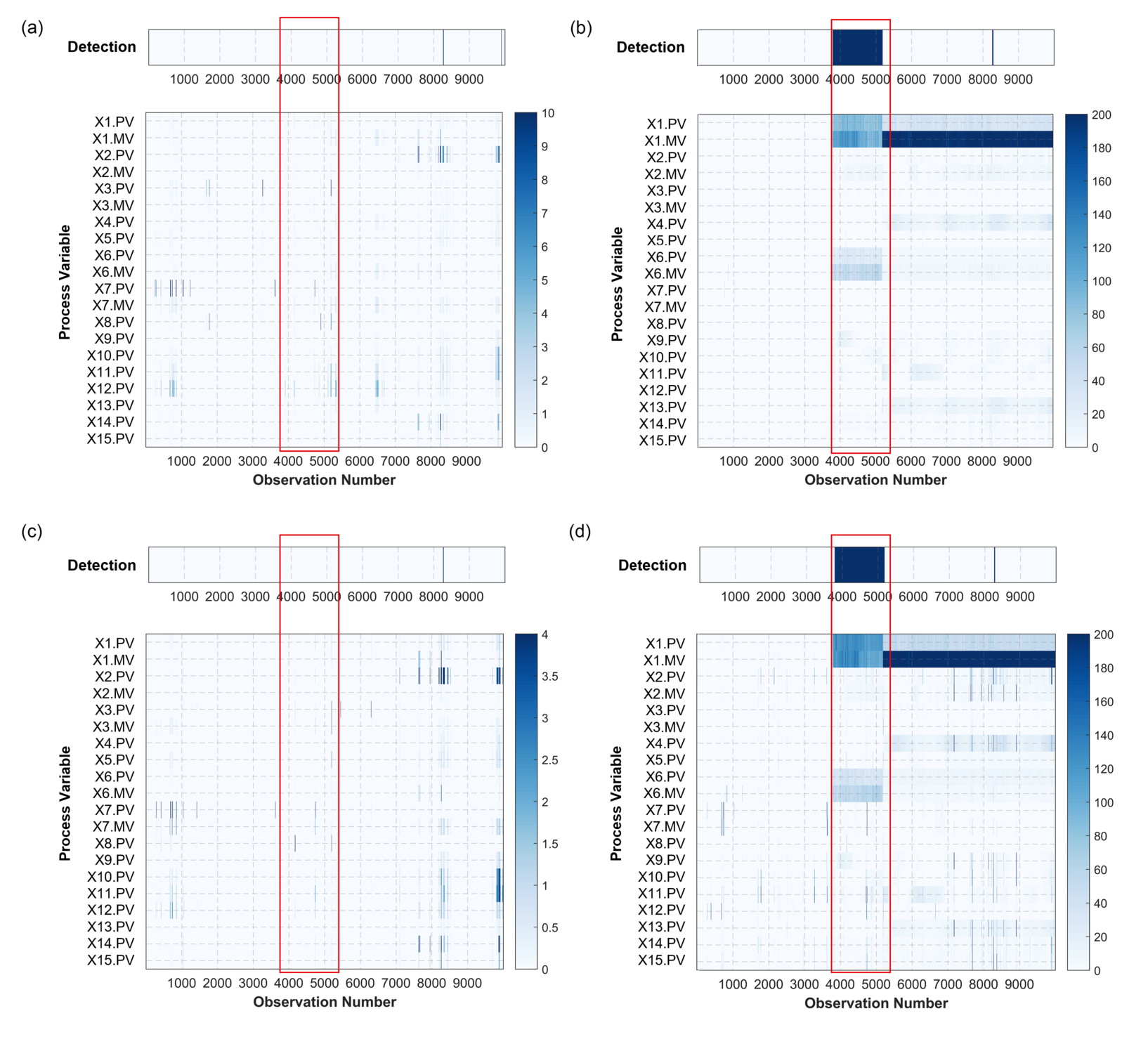}
  \caption{FDI by (D)PCA methods for Fault 1: (a) r-PCA, (b) f-PCA, (c) r-DPCA, and (d) f-DPCA.
    The red box indicates the period with the foaming event recorded by the operator.}
  \label{fig:fault1:pca}
\end{figure}

For comparison, the corresponding FDI results by (D)PCA methods are shown in Figure \ref{fig:fault1:pca}. The r-PCA and r-DPCA models simply fail to detect the fault. In the contribution plots, the r-PCA and r-DPCA models also fail to identify any variable that is noticeably affected by the foaming event. While (D)PCA models with reduced dimensionality determined by parallel analysis have been widely applied \citep{Chiang2000,Yin2014,De2015,Valle1999}, they are incapable of accurately detecting the main fault in this case.
For f-PCA and f-DPCA models, the $T^2$ statistic is able to detect the documented fault. The contribution plots also identify X1.PV and X6.PV as being associated with the fault. However, X1.MV and X6.MV are also identified as abnormal and as more significantly than X1.PV and X6.PV. While those variables are likely affected by the fault, they are operating normally with respect to the control system dynamics and thus should not have been identified. Furthermore, some of these variables continue to be highlighted well after the issue is resolved. The (D)PCA models also only scantly detect the deviations in the later time that are correctly highlighted by the BRNN.

\begin{figure}
  \centering
  \includegraphics[width=0.48\textwidth]{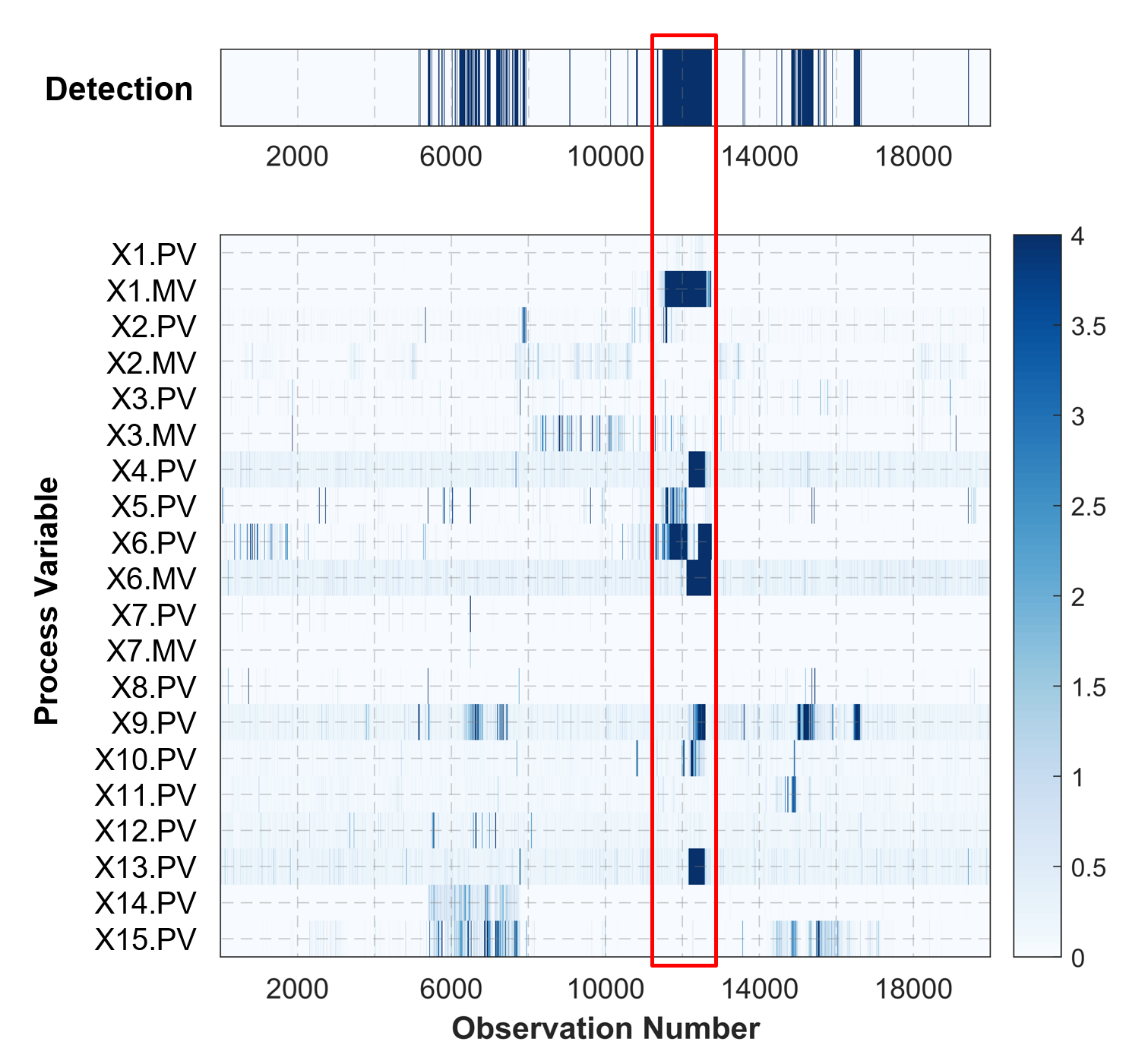}
  \caption{FDI by BRNN for Fault 2.
    The red box indicates the period with the foaming issue as recorded by the operator.}
  \label{fig:fault2:brnn}
\end{figure}

\subsubsection{Fault 2 results}%
\label{sec:real-data:fault2}

The BRNN FDI results for Fault 2 are shown in Figure \ref{fig:fault2:brnn}. The proposed method successfully detects the documented fault and identifying the related variables, as marked by time interval with the red box. For the earlier period around the $6000^{\mathrm{th}}$ time point, the BRNN model detects a disturbance and identifies the deviation in X14.PV, X15.PV, and X9.PV. The relative magnitude of the deviation during those periods is not as significant as that during the documented fault period. This assessment was then verified to be fully warranted by inspection of the recorded sensor measurements. Analysis of the period around the $15000^{\mathrm{th}}$ data point yield similar results.

\begin{figure}
  \centering
  \includegraphics[width=0.96\textwidth]{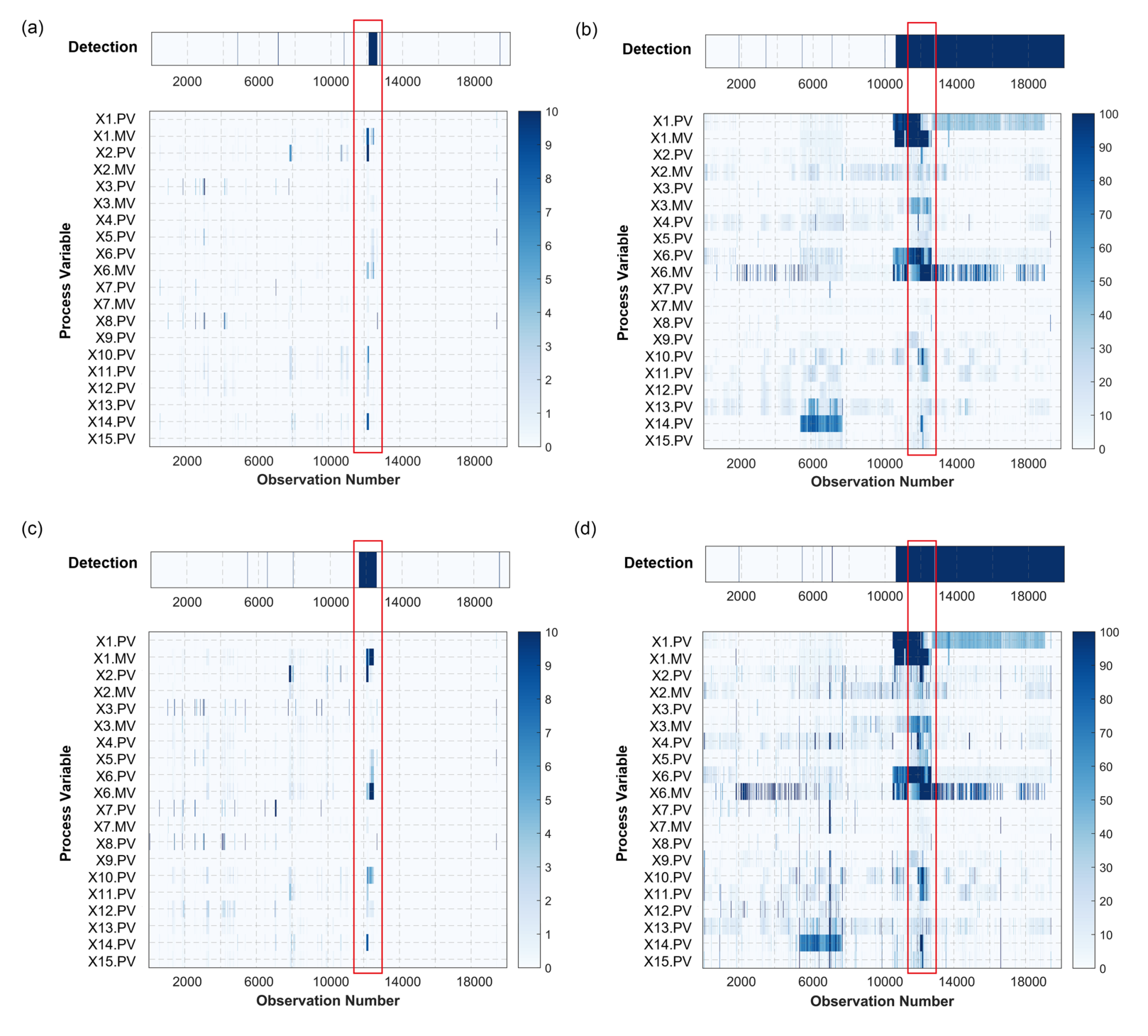}
  \caption{FDI by (D)PCA methods for Fault 2: (a) r-PCA, (b) f-PCA, (c) r-DPCA, and (d) f-DPCA.
    The red box indicates the period with the foaming issue as recorded by the operator.}
  \label{fig:fault2:pca}
\end{figure}

As before, the (D)PCA FDI methods are also applied. Their results are shown in Figure \ref{fig:fault2:pca}. As observed for Fault 1, the r-PCA and r-DPCA models are not as sensitive to the fault and only partially detect the documented fault period, and they did not detect the earlier event highlighted by the BRNN method. The identification plots by the r-PCA and r-DPCA models in Figures \ref{fig:fault2:pca}ac also only identify a limited number of faulty variables. The f-PCA and f-DPCA models are again overly sensitive for both FDI, flagging much of the data period. After the foaming issue occurred and the operator intervention, the control system is able to compensate for the disturbance after a while. However, f-PCA and f-DPCA models incorrectly continue to assess the system as in an abnormal state even though the foaming issue has been fully resolved. This can also be observed from the contribution plots in Figures \ref{fig:fault2:pca}bd, wherein variables X1.PV and X6.MV are identified as problematic during and long after the resolution of the fault.

To summarize, this case study on real data from a chemical process demonstrates the higher accuracy, specificity, and robustness in FDI of the BRNN-based method over (D)PCA-based methods. The proposed method is also shown to provide precise and easily interpretable results for prompt diagnosis and mitigation of fault events in real manufacturing processes.

\section{Conclusion}%
\label{sec:conclusion}

This article proposes a novel BRNN-based FDI method for manufacturing processes. 
The proposed method simultaneously tackles three key challenges in modeling real process data: (1) concurrent spatio-temporal correlations, (2) nonlinearity, and (3) incomplete characterization of the uncertainty in process noise and dynamics.
The BRNN model addresses these challenges because of its probabilistic framework built on RNN models. And, for implementation efficiency, the inference is made using variational dropout, which both regularizes the NN during training and efficiently estimates the uncertainty as it evolves through time.

The uncertainty estimates of the BRNN model play a crucial role in FDI. By continuously estimating the uncertainty, the BRNN model provides adaptive confidence intervals that fully characterize the system dynamics based on the current and past information. As demonstrated here, the BRNN framework therefore enables:
\begin{enumerate}[(1)]
  \item fault detection in processes with nonlinear dynamics, and
  \item direct fault identification with easily interpreted identification plots and fault 
    propagation path analysis.
\end{enumerate}

The effectiveness of the proposed BRNN method is demonstrated in two case studies: (1)  the benchmark TEP dataset and (2) a real chemical manufacturing dataset. The proposed method is compared to the widely applied PCA and DPCA methods, using either full and reduced dimension models. The comparisons show that the BRNN model provides results that are accurate and more specific and directly relevant for fault identification. Furthermore, based on its results, one can distinguish the nature of the faults, between controllable, back to control, or uncontrollable faults.

More broadly, the application of Bayesian methods to fault detection is not a widely explored field. To that end, this paper demonstrates a novel framework involving the systematic application of spatio-temporal models with Bayesian estimation such that the posterior inference results are directly relevant for detection and identification. In this case, a BRNN is used, but the strategy could be adapted for other spatio-temporal models such as dynamic process models.

The proposed BRNN-based FDI framework can be directly applied to any manufacturing process with historical NOC measurements without significant modifications. Moreover, the easy implementation of variational dropout to any model architecture and concurrent online calculating capability make BRNN feasible for large-scale industrial applications.

Some considerations for future work might include:
\begin{enumerate}[(1)]
  \item The online adaptation of the BRNN model for changing NOC. In real chemical processes, the process conditions evolve and it is unlikely that the training data can cover all of the NOC modes. Thus, online adaptation is crucial for reducing false alarms and maintenance costs.

  \item While proposed here for FDI, the BRNN model framework also has broad potential applications in industrial manufacturing processes related to time series analysis. The variational dropout can be applied to any deep learning model without modification of the model architecture, which makes it a preferable probabilistic model as compared to other recent advanced techniques.
\end{enumerate}

\newcommand{\noopsort}[1]{} \newcommand{\printfirst}[2]{#1}
  \newcommand{\singleletter}[1]{#1} \newcommand{\switchargs}[2]{#2#1}

\end{document}